\documentclass[letterpaper]{article} 
\usepackage[preprint]{paper2027}   
\usepackage[hyphens]{url}  
\usepackage{graphicx} 
\usepackage{natbib}  
\usepackage{caption} 
\usepackage{xcolor} 
\usepackage{multirow} 
\usepackage{adjustbox} 
\usepackage{amsmath} 
\usepackage{amssymb} 
\usepackage{algorithm}
\usepackage{algorithmic}

\usepackage{newfloat}
\usepackage{listings}
\DeclareCaptionStyle{ruled}{labelfont=normalfont,labelsep=colon,strut=off} 
\floatstyle{ruled}
\newfloat{listing}{tb}{lst}{}
\floatname{listing}{Listing}

\usepackage{booktabs}
\usepackage{dblfloatfix}

\title{Grounding and Explaining Visual Evidence for AI-Generated Image Detection in Human-Centric Scenes}
\author{
    Kun Guo\textsuperscript{\rm 1},
    Yuzhou Yang\textsuperscript{\rm 1},
    HaoYue Wang\textsuperscript{\rm 1},
    Qichao Ying\textsuperscript{\rm 2},
    Sheng Li\textsuperscript{\rm 1},
    Zhenxing Qian\textsuperscript{\rm 1}\corresponding}
\affiliations{
    \textsuperscript{\rm 1}College of Computer Science and Artificial Intelligence, Fudan University\\

    \textsuperscript{\rm 2}NVIDIA
   
}

\begin{document}
\maketitle


\begin{abstract}
  Rapid advances in image generation models call for interpretable AI-generated image detection methods that not only determine authenticity but also provide supporting visual evidence. Existing approaches may produce inconsistencies between generated explanations and localized evidence regions, undermining the reliability of explanations for authenticity decisions. Meanwhile, existing benchmarks provide limited coverage of the diverse human-centric scenes prevalent in generated imagery. To address these limitations, we investigate authenticity detection with grounded and explainable visual evidence in human-centric scenes. We present HAVE (Human-centric AI-generated Visual Evidence), a diverse human-centric dataset comprising 40K real and 39K AI-generated images from 10 recent generators, with 106K localized evidence instances across 8 evidence categories, each annotated with a bounding box and a region-aligned explanation. We further propose PAVE, a Perception-Aware Visual Evidence framework that jointly performs authenticity prediction, visual evidence grounding, and region-aligned explanation generation. PAVE employs a judge-guided alignment reward to assess region--explanation consistency and evidence validity, together with perception-aware regularization that contrasts token-level predictions between original and randomly masked images to promote reliance on visual input. Experiments on HAVE and external datasets demonstrate strong performance in authenticity detection, visual evidence grounding, and explanation quality. Code will be public at https://github.com/guokun111/PAVE.
\end{abstract}


\section{Introduction}


\begin{table*}[!t]
  
  \tabcolsep=9 pt
  \centering
  \begin{adjustbox}{width=\linewidth}
  \begin{tabular}{r|l|rr|c|rr|rr|rr}
  \toprule
  \multirow{2}{*}{Dataset} & \multirow{2}{*}{Venue} & \multicolumn{2}{c|}{Scale} & \multirow{2}{*}{Image Size} & \multicolumn{2}{c|}{Prompt} & \multicolumn{2}{c|}{Generator Count} & \multicolumn{2}{c}{Evidence Annotation} \\
   & & Real & Gen. & & Avg. Len. & Struct. & 2025 & 2026 & Category & Instance \\
  \midrule
  
  PAL4VST \cite{zhang2023perceptual}  & ICCV  & 0  & 10168  & -  & 5   & $\times$  & -  & -  & -  & 40,841  \\
  RichHF-18K \cite{liang2024rich}     & CVPR  & 0  & 11140  & 512$\times$512-768$\times$768  & -   & $\times$  & -  & -  & -  & 82,430  \\
  LOKI \cite{ye2025loki}              & ICLR  & 900  & 229  & 512$\times$512-1024$\times$1331  & -   & $\times$  & 0  & -  & -  & 687  \\
  SythScars \cite{kang2025legion}     & ICCV  & 0  & 12236  & 224$\times$224-768$\times$768  & -   & $\times$  & 0  & -  & 13  & 26,566  \\
  X-AIGD \cite{X-AIGD}                & ICLR  & 3868  & 3337  & 624$\times$416-1248$\times$1024  & 77  & $\times$  & 0  & 0  & 7  & 18,285  \\
  Fakexplain\cite{ji2026fakexplain}   & ICLR  & 8772  & 8772  & 512$\times$512-1024$\times$1024  & 5   & $\times$  & 0  & 0  & 6  & 31,395  \\
  \midrule
  
  \textbf{HAVE (Ours)}  & \multicolumn{1}{c}{-}  & 40000  & 39459  & 1152$\times$864-5504$\times$3072  & 147  & $\checkmark$  & 6  & 4  & 8  & 106,609 \\
  \bottomrule
  \end{tabular}
  \end{adjustbox}
  \caption{Comparison with existing AI-generated image datasets with evidence annotation. Gen., Avg. Len., and Struct. denote generated images, average prompt length, and structured prompts, respectively.}
  \label{tab:dataset_comparison}
\end{table*}

Rapid advances in image generation models~\cite{fan2025fluidauto2,tian2024visualauto1} have enabled the creation of photorealistic images, substantially lowering the barriers to content production and facilitating creative  expression. However, the misuse of such technologies poses risks of  misinformation dissemination~\cite{xu2023combatingmisinformation}. Among various types of generated content, human-centric images are especially prevalent~\cite{yang2026human} and can cause more immediate and consequential societal harm, as they may be misused to fabricate identities and events, manipulate public opinion, and undermine social ethics~\cite{momeni2025political, kaushik2025fraud}. Therefore, developing reliable approaches for detecting AI-generated images in human-centric scenes has become increasingly urgent.

 Early AI-generated image detection methods~\cite{tan2024npr,wang2020cnnspot,ojha2023towards} typically formulate the task as image-level binary classification. Recent efforts~\cite{wen2026fakeclue,gao2026toward} leverage Vision-Language Models (VLMs) to improve detection accuracy while providing human-readable explanations. However, these methods rely heavily on prompt engineering to obtain model-generated explanations, which are prone to hallucinations and lack spatial alignment with specific image regions~\cite{ji2026fakexplain}. Some studies~\cite{kang2025legion,xu2025fakeshield} use such textual explanations to guide auxiliary segmentation modules (e.g., SAM~\cite{kirillov2023sam}) in grounding suspicious evidence regions. Nevertheless, inconsistencies between the explanations and the actual evidence regions may compromise grounding accuracy. Moreover, such approaches tend to focus on semantic objects within the image rather than the visual evidence that supports the authenticity judgment.
 
Alongside advances in detection methods, numerous AI-generated image datasets have been introduced. Most of the early work~\cite{xiao2025AIGIDataset,yan2025Chameleon} provide only binary classification labelsyet lack human-readable explanations, limiting their applicability to interpretable AI-generated image detection. Some datasets~\cite{wen2026fakeclue,gao2026toward} provide textual explanations but do not annotate the corresponding image regions, lacking spatial alignment between the explanations and the visual content. More recent datasets~\cite{X-AIGD,ji2026fakexplain} provide both grounded evidence regions and their associated explanations; however, they primarily target relatively simple natural scenes with limited compositional and categorical diversity, overlooking the complex human-centric scenarios prevalent in generated imagery. This limitation restricts their applicability to diverse real-world settings. 

To address these challenges, we first construct HAVE, a dataset of Human-centric AI-generated Visual Evidence covering realistic and diverse everyday scenes. We derive structured visual descriptions from human-centric photographs and synthesize images with 10 recent generators under diverse generation settings, using each description to generate only one image. Furthermore, HAVE provides fine-grained visual evidence annotations consisting of bounding boxes, evidence categories, and region-aligned explanations. We employ an annotation refinement process to ensure the spatial and semantic alignment of each localized region with its corresponding explanation.
Furthermore, we introduce PAVE, a Perception-Aware Visual Evidence framework for grounding and explaining visual evidence. To align visual evidence with textual explanations, we design a judge-guided alignment reward that evaluates whether each region-aligned explanation is consistent with its corresponding visual region and whether each predicted bounding box contains valid visual evidence. Moreover, we introduce perception-aware regularization that compares token-level generation probabilities between the original input image and a randomly patch-masked version, encouraging divergent predictions when the visual evidence is perturbed. This allows the model to ground its responses in image content, enhancing visual evidence grounding performance and explanation quality.

Our main contributions are summarized as follows:
\begin{itemize}
\item We construct \textbf{HAVE}, a large-scale human-centric benchmark for AI-generated image detection that provides fine-grained visual evidence annotations with region-aligned explanations.

\item We propose \textbf{PAVE}, a reliable framework that jointly performs authenticity prediction, visual evidence grounding, and region-aligned explanation generation, helping users detect AI-generated images and understand the underlying visual evidence.

\item We introduce a devised judge-guided alignment reward and perception-aware regularization, enforcing that predictions are grounded in visible image content and aligned with corresponding evidence regions.

\item  Extensive experiments demonstrate the effectiveness of our method in authenticity detection, visual evidence grounding, and explanation quality. Our method also exhibits robust generalization across diverse datasets.

\end{itemize}

\section{Related Work}
\begin{figure*}[!t]
  \centering
  \includegraphics[width=\textwidth]{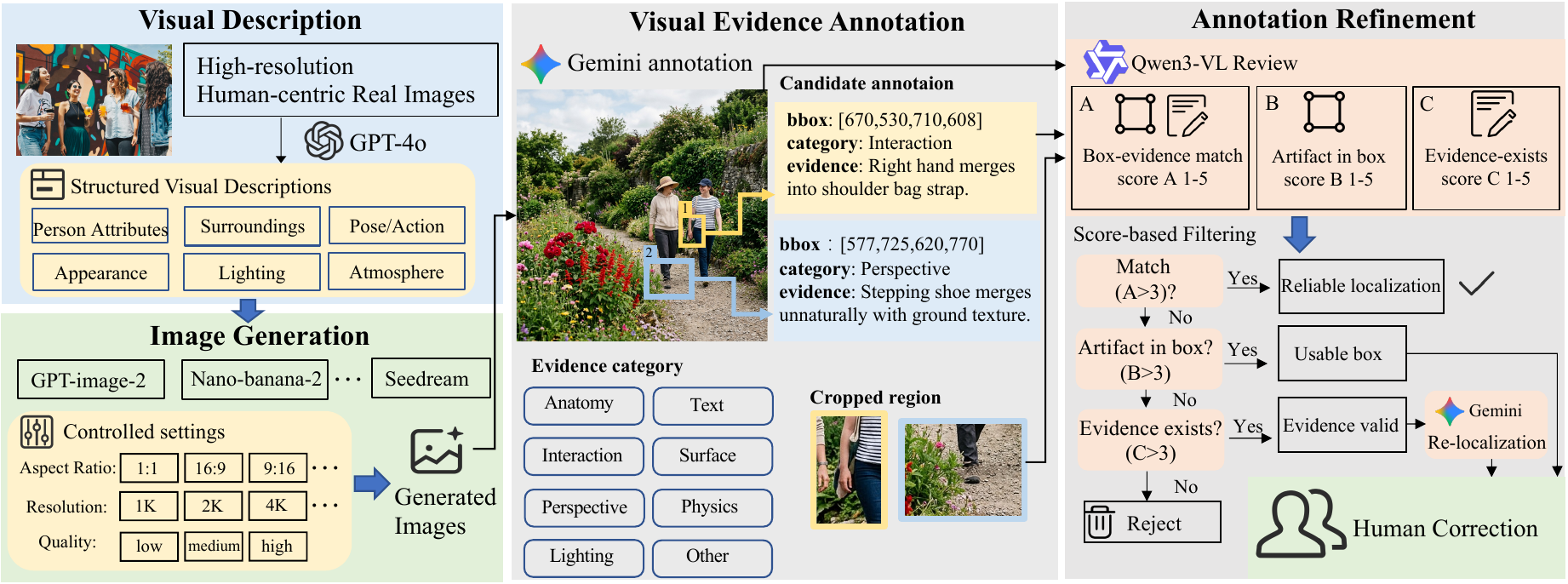}
  \caption{Overview of the HAVE construction pipeline. High-resolution human-centric images are converted into structured visual descriptions and synthesized with recent generators under diverse settings. Candidate evidence annotations are then reviewed, filtered, relocalized when necessary, and manually corrected.}
  \label{fig:dataset_framework}
  \end{figure*}
  
\begin{figure}[!t]
  \centering
  \includegraphics[width=\columnwidth]{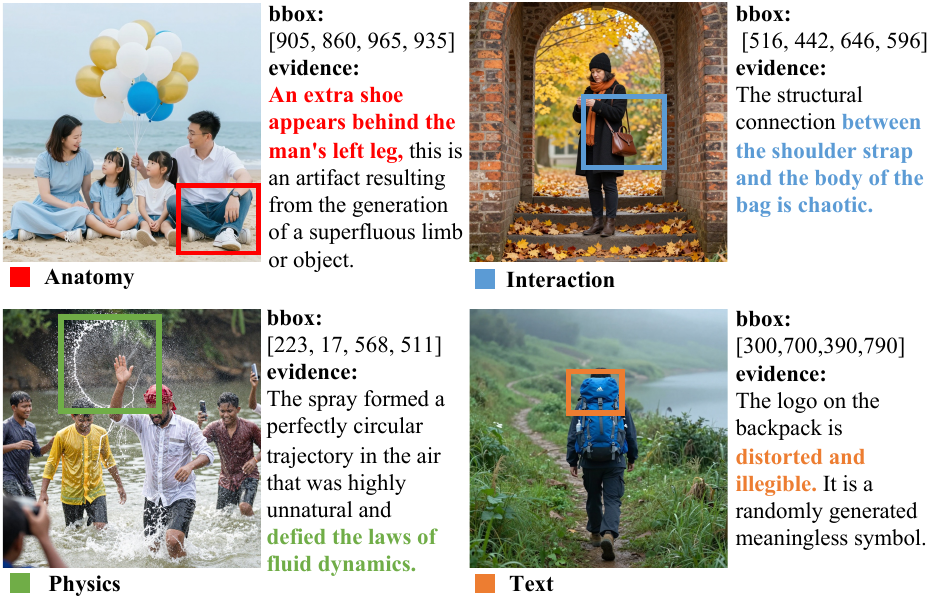}
  \caption{Examples of four visual evidence categories in HAVE. Colored boxes indicate evidence regions, accompanied by normalized coordinates and concise explanations.}
  \label{fig:dataset_showcase}
\end{figure}
\subsection{AI-Generated Image Datasets}

Existing benchmarks~\cite{zhu2023genimage,xiao2025AIGIDataset} primarily support image-level classification. Recent datasets~\cite{wen2026fakeclue,gao2026toward} provide  explanations. However, they lack spatial localization annotations and do not explicitly align explanations with their supporting visual regions. Several datasets provide spatial annotations for generated-image artifacts. PAL4VST~\cite{zhang2023perceptual} and RichHF-18K~\cite{liang2024rich} localize perceptual defects with masks or points, but they target generation quality assessment rather than AI-generated image detection and lack region-aligned explanations. LOKI~\cite{ye2025loki}, SynthScars~\cite{kang2025legion}, X-AIGD~\cite{X-AIGD}, and FakeXplained~\cite{ji2026fakexplain} provide bounding boxes or masks together with evidence categories or descriptions. However, they focus on relatively simple scenes or older generators with more obvious artifacts. Such datasets therefore provide limited support for jointly capturing diverse human-centric scenes and offering fine-grained annotations that align suspicious regions with categories and textual evidence.

\subsection{AI-Generated Image Detection Methods}
Early AI‑generated image detectors formulate the task as binary classification. Some methods ~\cite{wang2020cnnspot,tan2024npr,chen2024drct} achieve effective classification, yet they provide little human‑interpretable evidence. To improve interpretability, recent MLLM-based methods~\cite{wen2026fakeclue,gao2026toward,zhou2025holmes}generate explanations together with authenticity predictions. However, these explanations lack explicit correspondence with suspicious regions. FakeShield~\cite{xu2025fakeshield} relies on an external segmentation module to localize manipulated regions. LEGION~\cite{kang2025legion} couples an MLLM with a grounding encoder and a pixel decoder, yet optimizes its classifier and localization branch as separate components. FakeXplainer~\cite{ji2026fakexplain} jointly produces verdicts, bounding boxes, and explanations through staged reinforcement learning, while Locate-Then-Examine~\cite{ji2026lte} re-examines cropped regions at the cost of multi-stage inference. 
Nonetheless, these methods do not ensure that the authenticity decision genuinely depends on visual perception of the localized evidence, which limits their visual evidence localization and explanation quality.

\section{HAVE Dataset}
As shown in Figure~\ref{fig:dataset_framework}, the HAVE construction pipeline comprises four stages: visual description, image generation, visual evidence annotation, and annotation refinement. We first derive structured descriptions from human-centric photographs and synthesize images using recent generators. We then annotate visual evidence with bounding boxes, categories, and explanations, followed by refinement to ensure annotation quality and region--explanation alignment.
\begin{figure*}[t]
  \centering
  \includegraphics[width=\linewidth]{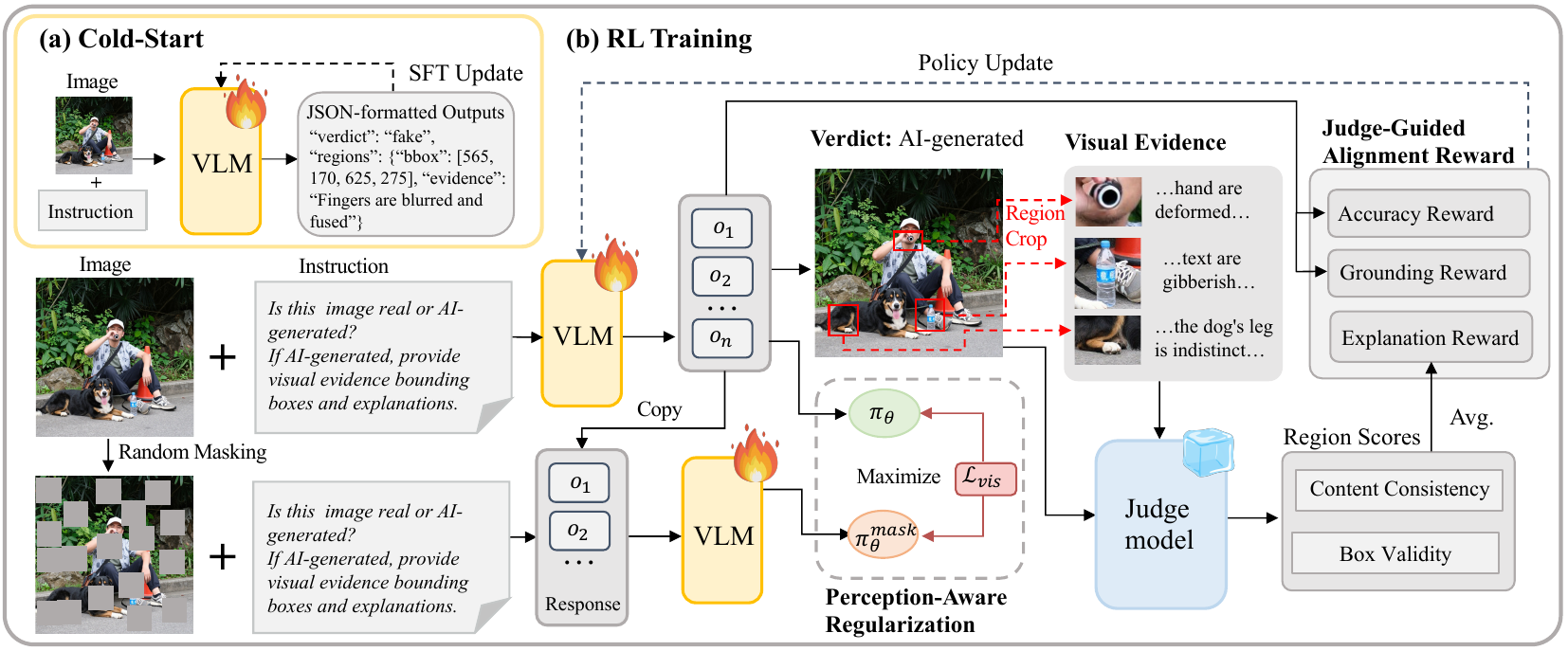}
  \caption{Overview of PAVE. (a) Cold-start SFT teaches the VLM to generate structured authenticity verdicts, evidence boxes, and region-aligned explanations. (b) RL training employs a judge-guided alignment reward to evaluate prediction accuracy, evidence grounding, and region--explanation consistency. Perception-aware regularization further promotes reliance on visual evidence by maximizing the output discrepancy between original and randomly masked images.}
  \label{fig:method_overview}
\end{figure*}
\subsection{Construction Workflow}
\noindent\textbf{Visual Description.}
This stage constructs structured visual descriptions from high-resolution human-centric images collected from PortraitCraft~\cite{sha2026portraitcraft}, a dataset sourced from Unsplash and covering diverse daily scenarios. To ensure that the generated images closely reflect real-world scenes and capture comprehensive visual elements, we use GPT-4o with a structured prompt that requests a visual description from six semantic aspects: person, appearance, pose/action, surroundings, lighting, and atmosphere. The resulting descriptions contain about 150 tokens on average, providing detailed generation inputs while reducing the chance of omitting important scene information.

\noindent\textbf{Image Generation.}
This stage synthesizes AI-generated images under diverse settings. Specifically, we use 10 commercial and open-source generators released in 2025–2026, including NanoBanana-2~\cite{nano2}, GPT-Image-1~\cite{gpt-image-1}, GPT-Image-2~\cite{gpt-image-2}, NanoBanana-1~\cite{nano1}, Qwen-Z~\cite{team2025zimage}, Seedream-3.0~\cite{gao2025seedream}, Seedream-4.0~\cite{seedream2025seedream}, Seedream-4.5~\cite{seedream45}, Wan-2.6~\cite{Wan2.6-image}, and Wan-2.7~\cite{Wan2.7-image}, abbreviated as Banana2, GPT-1, GPT-2, Banana1, Qwen-Z, SeeD-3.0, SeeD-4.0, SeeD-4.5, Wan-2.6, and Wan-2.7, respectively.
To diversify generation settings, we vary supported aspect ratios, resolutions, and quality levels, including 1:1 and 9:16 ratios and resolutions from approximately 1K (1152×864) to near 4K (5504×3072). 
In addition, each structured visual description is assigned to only one generator to produce one image. This prevents the same scene description from appearing in multiple generator subsets, reducing the risk that models exploit repeated semantic compositions or layouts rather than learning genuine generation evidence.

\noindent\textbf{Visual Evidence Annotation.}
This stage annotates the generated images with visual evidence. We first use Gemini-3.1-pro to propose candidate evidence regions in each image. As illustrated in Figure~1, each candidate annotation consists of three components: (1) a bounding box that localizes the visible evidence region with surrounding context, (2) an evidence category that specifies the type of the observed visual evidence, and (3) a textual evidence note that provides an explanation of the visible abnormality within the annotated region. Each visual evidence instance corresponds to one of eight evidence categories, including Anatomy, Text, Perspective, Interaction, Surface, Lighting, Physical, and Other. This design focuses on where the visual evidence appears and what it supports for the authenticity judgment.

\noindent\textbf{Annotation Refinement.}
This stage further reviews the candidate evidence annotations through an auditable annotation pipeline. Specifically, for each candidate annotation, we feed the full image with the annotated bounding box, the cropped region of the annotation, and the corresponding textual evidence into Qwen3-VL, a powerful visual understanding model. It then scores each candidate from three complementary aspects: whether the boxed region contains visible visual evidence, whether the textual evidence is visually grounded in the image, and whether the box and the evidence refer to the same visual evidence. Each aspect is scored on a 1-to-5 scale, where higher scores indicate higher annotation reliability.
The resulting scores are used for score-based filtering, which assigns each candidate box to a single outcome. Candidates with consistent box-evidence alignment are kept as reliable annotations. Candidates whose boxes contain visible visual evidence are also retained as usable localized evidence, while their textual evidence is manually reviewed and corrected to align with the retained box. Candidates with valid evidence but misaligned boxes are sent to Gemini for evidence-guided bbox re-localization, followed by manual inspection and correction using our custom refinement tool. Those without reliable visual evidence are discarded. This process results in refined evidence annotations with localized regions, evidence categories, and textual evidence.

\subsection{Dataset Statistics and Showcase}
As summarized in Table~\ref{tab:dataset_comparison}, our dataset exhibits advantages over existing datasets in terms of image resolution, annotation scale, recent-generator coverage, and prompt richness. Figure~\ref{fig:dataset_showcase} presents representative evidence samples from HAVE to illustrate the diversity and granularity of our annotations. We divide HAVE into training and test sets at a 3:1 ratio. Specifically, the training set contains 28,615 generated images and 30,000 real images, while the test set contains 10,844 generated images and 10,000 real images. Detailed distributions of generators, evidence categories, and annotation outcomes are provided in the appendix.

\section{The Proposed PAVE}

As shown in Figure~\ref{fig:method_overview}, PAVE is first trained through supervised fine-tuning (SFT) to establish a stable foundation for generating the required structured responses. We then perform reinforcement learning (RL) using a judge-guided alignment reward and perception-aware regularization to improve the consistency between grounded evidence regions and their corresponding explanations while encouraging reliance on visual content.
\subsection{Task Formulation and Cold Start}
Given an input image $I$ and a task instruction $p$, the VLM autoregressively generates a JSON-formatted structured response $y$ as
\begin{equation}
y = \{v, E\},
\qquad
E = \{(b_k,e_k)\}_{k=1}^{K}.
\end{equation}
where $v \in \{\text{real},\text{AI-generated}\}$ denotes the authenticity prediction and $E$ denotes the set of localized visual evidence. For the $k$-th evidence instance, $b_k=[x_1,y_1,x_2,y_2]$ is a normalized bounding box in a fixed coordinate space, and $e_k$ is a region-aligned explanation describing why the localized region supports the prediction.

For an AI-generated image, the target output contains an AI-generated verdict and one or more evidence instances. For a real image, the target output contains a real verdict and an empty evidence set, i.e., $E=\emptyset$. This formulation unifies authenticity prediction, visual evidence grounding, and region-aligned explanation within the same generation space, while discouraging false evidence on real images.

Before reinforcement learning, we perform full SFT of the VLM on HAVE dataset. Each training sample consists of an image--instruction pair $(I,p)$ and its ground-truth JSON-formatted response $y^{*}$. We serialize $y^{*}$ into a token sequence $(y_1^{*},\ldots,y_T^{*})$ and optimize the model by minimizing the token-level negative log-likelihood:
\begin{equation}
\mathcal{L}_{\mathrm{SFT}}
=
-\frac{1}{T}\sum_{t=1}^{T}
\log \pi_\theta
\bigl(y_t^{*}\mid I,p,y_{<t}^{*}\bigr).
\end{equation}
This stage teaches the VLM to follow the task instruction and produce valid structured responses in the required JSON format, thereby providing a stable initialization for subsequent reinforcement learning.

\subsection{Judge-Guided Alignment Reward}
Following SFT, we optimize the model through two successive reinforcement learning stages, each targeting a distinct aspect of model performance.

\noindent\textbf{Evidence Grounding.}
The first stage focuses on authenticity prediction and visual evidence grounding, encouraging the model not only to produce the correct authenticity verdict but also to localize the visible evidence supporting that verdict. Let $B=\{b_k\}_{k=1}^{K}$ and $B^{*}$ denote the predicted and ground-truth evidence regions, respectively. We measure their spatial agreement using the localization reward
\begin{equation}
R_{\mathrm{loc}}
=
\operatorname{IoU}(B,B^{*}),
\end{equation}
where the IoU is computed between the foreground masks formed by the union of all boxes in each set. For a real image, $B^{*}=\emptyset$, and the localization reward is maximized only when the model predicts no evidence region. We define $R_{\mathrm{loc}}=1$ when both $B$ and $B^{*}$ are empty, and $R_{\mathrm{loc}}=0$ when exactly one of them is empty.

To preserve authenticity prediction performance, we further define a classification reward as
\begin{equation}
R_{\mathrm{cls}}
=
\mathbb{I}(v=v^{*}),
\end{equation}
where $v^{*}$ denotes the ground-truth authenticity label and $\mathbb{I}(\cdot)$ is the indicator function, which equals $1$ when its argument is true and $0$ otherwise. The overall reward for this stage is defined as
\begin{equation}
R_{\mathrm{evi}}
=
R_{\mathrm{loc}}^{\alpha}
+
\eta R_{\mathrm{cls}},
\end{equation}
where $\alpha$ emphasizes high-quality evidence grounding and $\eta$ controls the contribution of authenticity classification. This reward evaluates sampled responses according to both verdict correctness and spatial grounding quality.

\noindent\textbf{Judge-Guided Alignment.}
The second stage aligns each explanation $e_k$ with its predicted evidence region $b_k$. For every region, a multimodal judge $\phi_{\mathrm{judge}}$ assigns two scores on a $1$-to-$5$ scale: a content-consistency score $c_k$, measuring how accurately $e_k$ describes the visible content inside $b_k$, and a box-validity score $a_k$, measuring whether the boxed content constitutes genuine visual evidence of image generation. To provide sufficient visual detail, we feed the judge a global context image with numbered boxes marking the approximate region locations, together with an original-resolution crop of each region. We normalize each score to $[0,1]$, denoted $\hat{c}_k$ and $\hat{a}_k$, and take their average as the per-region reward; the explanation reward is then averaged over the evaluated regions, i.e.,
\begin{equation}
s_k = \frac{\hat{c}_k + \hat{a}_k}{2},
\qquad
R_{\mathrm{exp}} = \frac{1}{|K|}\sum_{k\in K} s_k \in [0,1],
\end{equation}
where $K$ indexes the evaluated regions, with regions beyond a fixed budget truncated. For AI-generated images, we combine this reward with the localization and classification rewards retained from the first stage:
\begin{equation}
R_{\mathrm{align}}
=
\lambda_{\mathrm{exp}}R_{\mathrm{exp}}
+
\lambda_{\mathrm{loc}}R_{\mathrm{loc}}
+
\lambda_{\mathrm{cls}}R_{\mathrm{cls}}.
\end{equation}
The dominant explanation term improves the consistency between explanations and localized visual evidence, while the remaining terms preserve localization and classification capabilities. For real images, we omit $R_{\mathrm{exp}}$ and retain only the classification reward and the no-evidence constraint encoded.

\subsection{Perception-Aware Regularization}

To encourage the model to base its predictions on visible image content, we introduce perception-aware regularization inspired by \cite{wang2026papo}. Given an input image $I$, we construct a corrupted view $\tilde{I}$ by randomly masking a subset of image patches. As illustrated in Figure~\ref{fig:method_overview}, $\pi_\theta$ and $\pi_\theta^{\mathrm{mask}}$ denote the output distributions of the same VLM conditioned on the original image $I$ and the masked image $\tilde{I}$, respectively. The two branches share the same model parameters $\theta$.

For the same output sequence $y$, we measure their token-level distributional discrepancy as
\begin{equation}
\begin{aligned}
\mathcal{L}_{\mathrm{vis}}
&=
\frac{1}{|\mathcal{C}|}
\sum_{t\in\mathcal{C}}
\left(e^{\delta_t}-\delta_t-1\right),\\
\delta_t
&=
\log \pi_{\theta}^{\mathrm{mask}}(y_t\mid\tilde{I},p)
-
\log \pi_{\theta}(y_t\mid I,p),
\end{aligned}
\end{equation}
where $\mathcal{C}$ denotes the set of output tokens and $p$ denotes the task instruction. A larger $\mathcal{L}_{\mathrm{vis}}$ indicates that the model is more sensitive to perturbations of the visual evidence.

In both reinforcement learning stages, we combine the GRPO policy objective with this regularizer:
\begin{equation}
\mathcal{L}
=
\mathcal{L}_{\mathrm{policy}}(R_s)
-
\lambda_{\mathrm{vis}}\mathcal{L}_{\mathrm{vis}},
\quad
R_s\in\{R_{\mathrm{evi}},R_{\mathrm{align}}\},
\end{equation}
where $R_s$ denotes the evidence grounding reward or the judge-guided alignment reward, and $\lambda_{\mathrm{vis}}$ controls the regularization strength. Minimizing this objective maximizes the prediction discrepancy when visual evidence is perturbed, discouraging the model from ignoring the input image.

\section{Experiments}

\begin{table*}[!t]
\tabcolsep=8 pt
\centering
\begin{adjustbox}{width=\textwidth}
\begin{tabular}{l|cccccccccc|c}
\toprule
Method & Banana2 & GPT-1 & GPT-2 & Banana & Qwen-Z & SeeD-3.0 & SeeD-4.0 & SeeD-4.5 & Wan-2.6 & Wan-2.7 & Avg. \\
\midrule
NPR           & 95.59 & 83.73 & 63.64 & \textbf{96.86} & 95.05 & 83.27 & 70.86 & 62.45 & \textbf{97.50} & 94.14 & 84.31 \\
RINE          & 94.27 & 93.45 & 92.36 & 92.36 & 93.50 & \underline{95.45} & \underline{94.77} & 86.91 & 88.44 & 90.27 & 92.18 \\
AIDE          & 94.60 & 93.60 & 93.40 & 94.10 & 94.70 & 93.60 & 91.60 & 85.80 & 90.10 & 91.40 & 92.30 \\
Effort        & 86.09 & 84.95 & 85.86 & 85.14 & 84.82 & 84.45 & 83.55 & 86.73 & 84.99 & 84.64 & 85.12 \\
DGS-Net       & \underline{95.80} & \underline{95.40} & \underline{93.50} & 95.90 & \textbf{96.10} & 93.50 & 92.00 & \underline{87.50} & 95.80 & \textbf{95.00} & \underline{94.05} \\
FakeVLM       & 74.59 & 71.27 & 66.95 & 82.73 & 71.86 & 77.23 & 58.86 & 56.14 & 76.57 & 75.68 & 71.19 \\
FakeReas.     & 68.45 & 70.50 & 72.95 & 67.86 & 65.18 & 76.86 & 64.41 & 57.09 & 64.15 & 68.64 & 67.56 \\
FakeShield    & 65.91 & 65.45 & 68.05 & 66.00 & 64.91 & 85.32 & 46.82 & 65.64 & 66.88 & 66.14 & 66.11 \\
\midrule
\textbf{PAVE(Ours)}        & \textbf{98.73} & \textbf{96.51} & \textbf{96.56} & \underline{96.06} & \underline{95.62} & \textbf{96.94} & \textbf{94.90} & \textbf{94.92} & \underline{96.09} & \underline{94.79} & \textbf{96.11} \\
\bottomrule
\end{tabular}
\end{adjustbox}
\caption{Authenticity detection performance (Acc, \%) of different methods on HAVE dataset.
The best and second-best results in each row are highlighted in \textbf{bold} and \underline{underlined}, respectively.
}
\label{tab:detection_acc_transposed}
\end{table*}

\begin{table*}[!t]
\tabcolsep=2 pt
\centering
\begin{adjustbox}{width=\linewidth}
\begin{tabular}{l|*{10}{cc|}cc}
\toprule
\multirow{2}{*}{Method} & \multicolumn{2}{c|}{Banana2} & \multicolumn{2}{c|}{GPT-1} & \multicolumn{2}{c|}{GPT-2} & \multicolumn{2}{c|}{Banana} & \multicolumn{2}{c|}{Qwen-Z} & \multicolumn{2}{c|}{SeeD-3.0} & \multicolumn{2}{c|}{SeeD-4.0} & \multicolumn{2}{c|}{SeeD-4.5} & \multicolumn{2}{c|}{Wan-2.6} & \multicolumn{2}{c|}{Wan-2.7} & \multicolumn{2}{c}{Avg.} \\
                        & IoU & F1 & IoU & F1 & IoU & F1 & IoU & F1 & IoU & F1 & IoU & F1 & IoU & F1 & IoU & F1 & IoU & F1 & IoU & F1 & IoU & F1 \\
\midrule
Qwen3-VL         & \underline{10.07} & \underline{16.67} & \underline{17.80} & \underline{26.65} & \underline{10.29} & \underline{16.69} & \underline{12.88} & \underline{20.64} & \underline{12.28} & \underline{19.30} & \underline{12.92} & \underline{20.65} & \underline{12.37} & \underline{19.95} & \underline{13.94} & \underline{22.21} & \underline{14.75} & \underline{23.17} & \underline{12.51} & \underline{20.13} & \underline{12.98} & \underline{20.61} \\
FRD-Net          & 5.43 & 9.65 & 9.12 & 15.60 & 6.15 & 10.71 & 7.30 & 12.67 & 6.29 & 11.01 & 7.61 & 13.41 & 6.85 & 11.94 & 7.76 & 13.33 & 8.23 & 14.09 & 7.61 & 13.13 & 7.23 & 12.55 \\
SparseViT        & 5.10 & 9.20 & 9.71 & 16.51 & 5.76 & 10.27 & 6.73 & 11.74 & 5.67 & 10.06 & 7.14 & 12.66 & 6.43 & 11.26 & 7.48 & 12.93 & 7.55 & 13.06 & 7.14 & 12.46 & 6.87 & 12.02 \\
FakeShield       & 4.04 & 7.01 & 6.75 & 11.48 & 4.61 & 7.80 & 5.09 & 8.62 & 3.89 & 6.84 & 5.50 & 9.49 & 4.97 & 8.52 & 4.98 & 8.57 & 5.80 & 9.95 & 5.02 & 8.59 & 5.07 & 8.69 \\
LEGION           & 2.00 & 3.40 & 4.39 & 7.13 & 2.97 & 5.01 & 3.18 & 5.27 & 2.55 & 4.14 & 4.24 & 7.01 & 2.63 & 4.30 & 3.40 & 5.52 & 4.38 & 7.01 & 3.65 & 5.93 & 3.34 & 5.47 \\
\midrule
\textbf{PAVE(Ours)}             & \textbf{27.06} & \textbf{38.27} & \textbf{38.23} & \textbf{50.71} & \textbf{26.03} & \textbf{36.74} & \textbf{33.52} & \textbf{45.41} & \textbf{30.47} & \textbf{42.02} & \textbf{37.53} & \textbf{50.29} & \textbf{26.88} & \textbf{37.86} & \textbf{27.48} & \textbf{38.43} & \textbf{33.16} & \textbf{44.75} & \textbf{30.73} & \textbf{42.13} & \textbf{31.11} & \textbf{42.66} \\
\bottomrule
\end{tabular}
\end{adjustbox}
\caption{Visual evidence grounding performance (IoU and F1, \%) on HAVE dataset. The best and second-best results in each column are highlighted in \textbf{bold} and \underline{underlined}, respectively.
}
\label{tab:localization_ours_v_transposed}
\end{table*}

\begin{table*}[!t]
\tabcolsep=2 pt
\centering
\begin{adjustbox}{width=\linewidth}
\begin{tabular}{l|*{10}{cc|}cc}
\toprule
\multirow{2}{*}{Method} & \multicolumn{2}{c|}{Banana2} & \multicolumn{2}{c|}{GPT-1} & \multicolumn{2}{c|}{GPT-2} & \multicolumn{2}{c|}{Banana} & \multicolumn{2}{c|}{Qwen-Z} & \multicolumn{2}{c|}{SeeD-3.0} & \multicolumn{2}{c|}{SeeD-4.0} & \multicolumn{2}{c|}{SeeD-4.5} & \multicolumn{2}{c|}{Wan-2.6} & \multicolumn{2}{c|}{Wan-2.7} & \multicolumn{2}{c}{Avg.} \\
                        & R-L & CSS & R-L & CSS & R-L & CSS & R-L & CSS & R-L & CSS & R-L & CSS & R-L & CSS & R-L & CSS & R-L & CSS & R-L & CSS & R-L & CSS \\
\midrule
Qwen3-VL         & \underline{14.96} & \underline{48.15} & \underline{14.86} & \underline{50.32} & \underline{13.29} & \underline{44.83} & \underline{14.93} & \underline{49.34} & \underline{14.69} & \underline{48.97} & \underline{15.34} & \underline{50.23} & \underline{13.30} & \underline{45.19} & \underline{14.74} & \underline{46.24} & \underline{14.68} & \underline{47.12} & \underline{14.96} & \underline{49.54} & \underline{14.85} & \underline{48.99} \\
FakeReas.        & 5.65 & 23.59 & 5.11 & 25.46 & 5.34 & 24.17 & 5.58 & 24.44 & 5.55 & 24.04 & 4.82 & 26.07 & 5.57 & 24.16 & 5.75 & 23.98 & 5.49 & 24.57 & 5.43 & 24.24 & 5.42 & 24.49 \\
FakeVLM          & 3.80 & 29.73 & 3.74 & 28.64 & 3.79 & 27.59 & 5.16 & 31.58 & 4.23 & 30.52 & 3.16 & 29.04 & 4.15 & 28.01 & 4.70 & 28.03 & 3.79 & 29.93 & 4.04 & 29.15 & 4.05 & 29.24 \\
FakeShield       & 4.80 & 31.18 & 5.38 & 33.05 & 4.38 & 29.72 & 5.19 & 32.17 & 4.53 & 31.63 & 5.62 & 33.14 & 4.27 & 30.78 & 4.47 & 31.49 & 4.99 & 31.80 & 4.88 & 31.12 & 4.97 & 31.80 \\
LEGION           & 7.58 & 44.26 & 8.87 & 47.85 & 6.71 & 41.76 & 8.57 & 46.51 & 7.59 & 45.57 & 9.36 & 48.66 & 7.03 & 43.34 & 7.09 & 43.60 & 7.69 & 45.14 & 7.66 & 44.64 & 7.81 & 45.13 \\

\textbf{PAVE(Ours)}             & \textbf{21.08} & \textbf{61.51} & \textbf{22.68} & \textbf{65.95} & \textbf{19.78} & \textbf{57.83} & \textbf{22.27} & \textbf{65.34} & \textbf{21.02} & \textbf{62.24} & \textbf{23.99} & \textbf{68.20} & \textbf{20.22} & \textbf{59.58} & \textbf{20.18} & \textbf{59.66} & \textbf{21.62} & \textbf{63.03} & \textbf{21.26} & \textbf{62.93} & \textbf{21.45} & \textbf{62.74} \\
\bottomrule
\end{tabular}
\end{adjustbox}
\caption{Explanation performance (ROUGE-L abbreviated as R-L and CSS, \%) on HAVE dataset. The best and second-best results in each column are highlighted in \textbf{bold} and \underline{underlined}, respectively.}
\label{tab:explanation_ours_v_transposed}
\end{table*}


\subsection{Experimental Settings}
\noindent\textbf{Implementation Details.} We adopt Qwen3-VL-Instruct 4B as our base model and train it on 8 NVIDIA H20 GPUs. In the SFT stage, we set the learning rate to $2\times10^{-5}$. In the first RL stage, we employ a learning rate of $1\times10^{-6}$ with a group size of $G=8$. The hyperparameters $\alpha$ and $\eta$ in Eq.~(5) are set to $1.1$ and $0.5$, respectively. In the second RL stage, we use a learning rate of $3\times10^{-7}$ with a group size of $G=4$, where Qwen3-VL-Instruct 32B serves as the frozen judge model. The reward weights $\lambda_{\mathrm{exp}}$, $\lambda_{\mathrm{loc}}$, and $\lambda_{\mathrm{cls}}$ in Eq.~(7) are set to $0.7$, $0.2$, and $0.1$, respectively.And $\lambda_{\mathrm{vis}}$ in Eq.~(9) to $0.01$. Each visual patch is independently masked with a probability of $0.6$. We utilize DeepSpeed ZeRO-2 for distributed training. 

\noindent\textbf{Baselines.}
For comparison, we re-train CNNSpot~\cite{wang2020cnnspot}, UnivFD~\cite{ojha2023towards}, NPR~\cite{tan2024npr}, RINE~\cite{koutlis2024RINE}, AIDE~\cite{yan2025Chameleon}, Effort~\cite{yan2025effort}, DGS-Net~\cite{yan2026dgsnet}, and FakeReasoning~\cite{gao2026toward} on the HAVE training set. For FakeVLM~\cite{wen2026fakeclue}, FakeShield~\cite{xu2025fakeshield}, LEGION~\cite{kang2025legion}, SparseViT~\cite{su2025sparsevit}, and FRD-Net~\cite{chen2026frd}, we adopt their officially released model weights and evaluation settings. We additionally evaluate three general-purpose open-source VLMs: InternVL2.5-26B~\cite{chen2024internvl}, Qwen3-VL-32B~\cite{bai2025qwen3}, and LLaMA3.2-Vision-11B~\cite{grattafiori2024llama}.


\noindent\textbf{Evaluation Metrics.}
Following prior works~\cite{xu2025fakeshield,kang2025legion}, we adopt their evaluation protocols and report binary classification accuracy for authenticity detection, the IoU and F1 score for visual evidence grounding, and ROUGE-L and the Cosine Similarity Score (CSS) for explanation quality. Higher values indicate better performance for all metrics.

\subsection{Experimental Results}
\begin{figure}[t]
  \centering
  \includegraphics[width=\columnwidth]{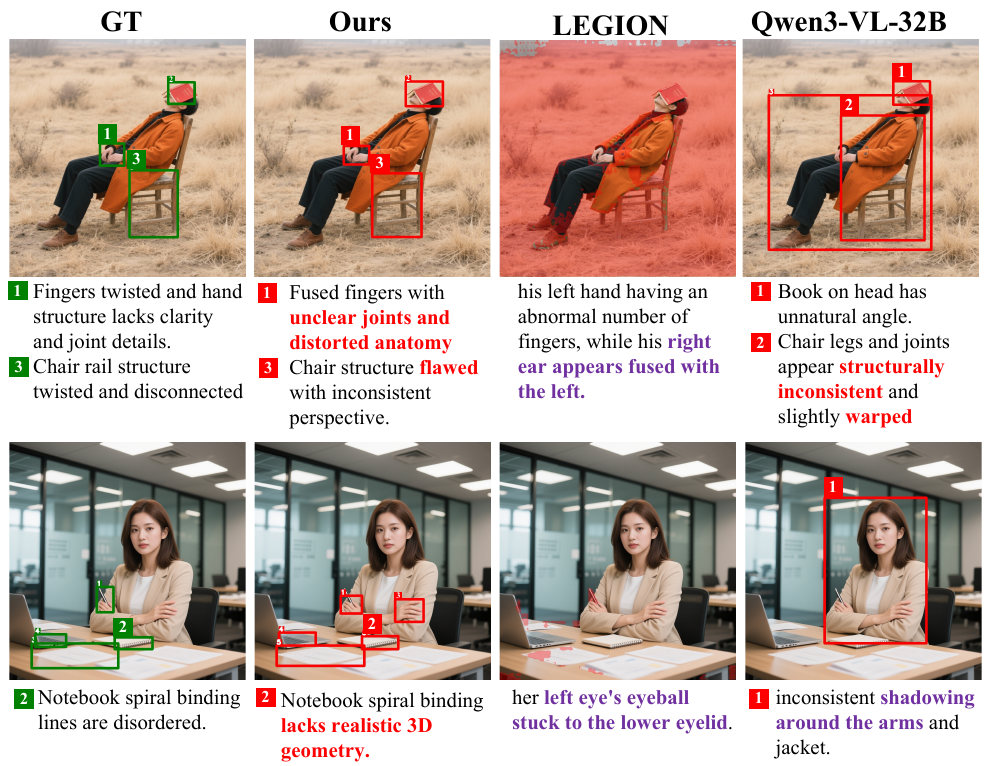}
  \caption{ Visual evidence grounding and explanation results of different models. Each numbered box is paired with its corresponding explanation. Explanations consistent with the ground-truth evidence are highlighted in red, whereas mismatched or unreliable explanations are shown in purple.}
  \label{fig:visualization}
\end{figure}

\noindent\textbf{Results on HAVE dataset.}  As shown in Table~\ref{tab:detection_acc_transposed}, PAVE achieves the highest average detection accuracy of 96.11\% across all ten generators, surpassing DGS-Net, the strongest retrained detector with 94.05\%, as well as larger VLM-based baselines. For visual evidence grounding, PAVE achieves an average IoU of 31.11\% and an F1 score of 42.66\% (Table~\ref{tab:localization_ours_v_transposed}), more than doubling the performance of the strongest baseline, Qwen3-VL, which attains 12.98\% IoU and 20.61\% F1. These gains highlight the difficulty of localizing reliable visual evidence in complex human-centric scenes and validate the motivation for constructing HAVE. For explanation quality, our method achieves the best CSS of 62.74\% and ROUGE-L of 21.45\% (Table~\ref{tab:explanation_ours_v_transposed}), outperforming Qwen3-VL by 13.75 and 6.60 percentage points, respectively. These consistent improvements demonstrate that our region-aligned objective effectively reduces the semantic gap between generated explanations and localized visual evidence.
Figure~\ref{fig:visualization} compares the grounding and explanation results of several methods, showing that PAVE accurately grounds visual evidence while generating explanations consistent with the localized regions.

\begin{table*}[!t]
\tabcolsep=16 pt
\centering
\begin{adjustbox}{width=\linewidth}
\begin{tabular}{l|rr|rr|rr|rr|rr}
\toprule
\multirow{2}{*}{Method}
  & \multicolumn{2}{c|}{SynthScars}
  & \multicolumn{2}{c|}{RichHF}
  & \multicolumn{2}{c|}{LOKI}
  & \multicolumn{2}{c|}{X-AIGD}
  & \multicolumn{2}{c}{Avg.} \\
& IoU & F1 & IoU & F1 & IoU & F1 & IoU & F1 & IoU & F1 \\
\midrule
Qwen3-VL & 9.89 & 14.76 & \underline{11.59} & \underline{19.28} & 11.65 & 16.29 & \underline{15.12} & \underline{15.86} & 12.06 & 16.55 \\
FakeShield & 7.66 & 11.54 & 6.07 & 10.78 & \underline{13.81} & \underline{21.55} & 6.23 & 9.49 & 8.44 & 13.34 \\
LEGION & \textbf{22.35} & \underline{32.23} & 11.41 & 19.23 & 9.87 & 15.99 & 9.32 & 14.47 & \underline{13.24} & \underline{20.48} \\
\midrule
\textbf{PAVE(Ours)} & \underline{22.20} & \textbf{33.20} & \textbf{19.12} & \textbf{30.70} & \textbf{17.93} & \textbf{26.95} & \textbf{16.91} & \textbf{25.64} & \textbf{19.04} & \textbf{29.12} \\
\bottomrule
\end{tabular}
\end{adjustbox}
\caption{Visual evidence grounding performance (IoU and F1, \%) of different methods across datasets.
The best and second-best results in each column are highlighted in \textbf{bold} and \underline{underlined}, respectively.
}
\label{tab:localization}
\end{table*}

\begin{table}[!t]
\tabcolsep=6 pt
\begin{adjustbox}{width=\linewidth}
\begin{tabular}{l|c|c|c|c|c}
\toprule
Method & Params & SynthScars & LOKI & X-AIGD & Avg. \\
\midrule
InternVL2.5 & 26B & 9.30 & 14.02 & 2.96 & 8.76 \\
LLaMA3.2-V & 11B & 11.17 & 16.21 & 6.58 & 11.32 \\
Qwen3-VL & 32B & 26.39 & 21.14 & 10.27 & 19.27 \\
FakeShield & 22B & 35.44 & 37.79 & 7.88  & 27.04 \\
LEGION     & 8B  & \textbf{53.49} & \underline{43.48} & \underline{11.74} & \textbf{36.24} \\
\midrule
\textbf{PAVE(Ours)}       & 4B  & \underline{48.42} & \textbf{44.19} & \textbf{14.47} & \underline{35.69} \\
\bottomrule
\end{tabular}
\end{adjustbox}
\caption{Explanation performance(CSS, \%) with model parameters of different methods.  The best and second-best results in each column are highlighted in \textbf{bold} and \underline{underlined}.}
\label{tab:explanation_method}
\end{table}

\noindent\textbf{Results on Cross-dataset.}
We evaluate PAVE and other VLM-based methods on four out-of-distribution datasets containing spatial annotations for AI-generated image artifacts, including SynthScars~\cite{kang2025legion}, LOKI~\cite{ye2025loki}, X-AIGD~\cite{X-AIGD}, and RichHF~\cite{liang2024rich}. The first three datasets provide paired artifact regions and textual explanations, whereas RichHF does not contain the region--explanation pairs required for explanation evaluation.

As shown in Table~\ref{tab:localization}, PAVE achieves the best overall visual evidence grounding performance, with an average IoU of 19.04\% and F1 score of 29.12\%, outperforming the second-best method by 5.80 and 8.64 percentage points, respectively. It obtains the highest F1 on all four datasets and the highest IoU on three. For explanation quality (Table~\ref{tab:explanation_method}), PAVE ranks first on LOKI and X-AIGD and achieves a competitive average CSS of 35.69\% using only 4B parameters. Notably, although SynthScars is an in-distribution test set for LEGION, PAVE achieves the second-best result on it and a comparable overall CSS to LEGION (35.69\% vs.\ 36.24\%), demonstrating strong cross-dataset generalization.

\begin{table}[t]
\centering
\setlength{\tabcolsep}{3pt}
\begin{adjustbox}{width=\columnwidth}
\begin{tabular}{cccc|ccccc}
\toprule
\multirow{2}{*}{SFT}
& Evi. & JG & PA
& \multirow{2}{*}{Acc$\uparrow$}
& \multirow{2}{*}{IoU$\uparrow$}
& \multirow{2}{*}{F1$\uparrow$}
& \multirow{2}{*}{R-L$\uparrow$}
& \multirow{2}{*}{CSS$\uparrow$} \\
& Ground. & Align. & Reg.
& & & & & \\
\midrule
$\checkmark$ & $\times$ & $\times$ & $\times$
& 0.936 & 0.181 & 0.269 & 0.177 & 0.519 \\

$\checkmark$ & $\checkmark$ & $\times$ & $\times$
& 0.914 & 0.222 & 0.315 & 0.177 & 0.526 \\

$\checkmark$ & $\checkmark$ & $\times$ & $\checkmark$
& 0.948 & 0.245 & 0.347 & 0.196 & 0.574 \\

$\checkmark$ & $\checkmark$ & $\checkmark$ & $\checkmark$
& \textbf{0.987} & \textbf{0.271} & \textbf{0.383}
& \textbf{0.211} & \textbf{0.615} \\
\bottomrule
\end{tabular}
\end{adjustbox}
\caption{Ablation study of different training stages in PAVE.
}
\label{tab:main_ablation}
\end{table}

\subsection{Ablation Studies.}
We conduct the ablation study on the Banana2 test subset of HAVE in Table~\ref{tab:main_ablation}. Starting from the SFT baseline, evidence-localization reinforcement learning improves IoU from 18.1\% to 22.2\% and F1 from 26.9\% to 31.5\%, although classification accuracy decreases slightly. Adding perception-aware regularization improves all five metrics and increases the accuracy to 94.8\%, indicating that encouraging reliance on visual input benefits detection, localization, and explanation generation. The explanation-alignment stage further improves ROUGE-L and CSS to 21.1\% and 61.5\%, respectively, while providing additional gains in classification and localization. Overall, the full model achieves the best performance across all metrics, demonstrating the complementary effects of the proposed design.

\section{Conclusion}

In this work, we investigate visual evidence grounding and explanation for AI-generated image detection in complex human-centric scenes. We construct HAVE, a benchmark comprising real images and AI-generated images produced by recent generators, with fine-grained annotations that align visual evidence regions with textual explanations. We propose PAVE, a unified VLM-based framework that jointly performs authenticity prediction, visual evidence grounding, and region-aligned explanation generation by employing a judge-guided alignment reward together with perception-aware regularization. Extensive experiments demonstrate that PAVE achieves superior performance while generalizing effectively across datasets.

\bibliography{references}

\appendix

\twocolumn[
  \begin{center}
    {\LARGE\bfseries Supplementary Materials}
  \end{center}
  \vspace{0.5\baselineskip}
]

\section{Related Work}
\label{sec:supp_related_work}

\begin{figure*}[!t]
\centering
\includegraphics[width=\textwidth]{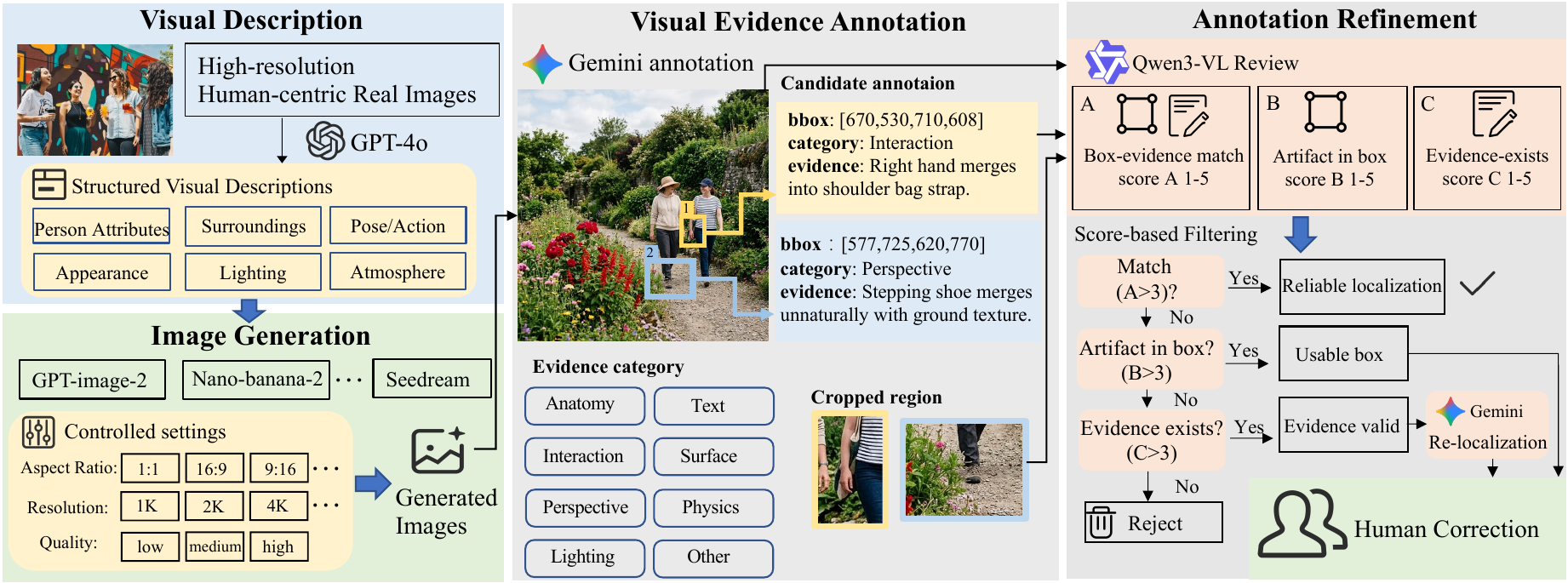}
\caption{Overview of the HAVE dataset construction pipeline. The pipeline comprises four stages: (1)~structured visual description from real human-centric photographs, (2)~synthetic image generation with recent generators under diverse settings, (3)~visual evidence annotation with bounding boxes and region-aligned explanations, and (4)~annotation refinement through score-based filtering and manual correction.}
\label{fig:supp_pipeline}
\end{figure*}

\subsection{AI-Generated Image Detection Methods}
\label{sec:supp_detection_methods}

Early AI-generated image detectors formulate the task as image-level binary classification. CNNSpot~\cite{wang2020cnnspot} enhances detector generalization via data augmentation. LGrad~\cite{tan2023LGrad} captures gradient-level traces left by image generators. NPR~\cite{tan2024npr} models neighboring pixel relationships to detect upsampling-induced artifacts. DIRE~\cite{wang2023dire} reconstructs images with a diffusion model and uses the reconstruction error as a detection signal. DRCT~\cite{chen2024drct} generates hard samples through diffusion reconstruction and applies contrastive learning to identify diffusion-specific traces. These methods achieve effective classification but provide little human-readable evidence for their decisions.

Recent MLLM-based methods generate natural-language explanations alongside authenticity predictions. FakeVLM~\cite{wen2026fakeclue} builds a dedicated MLLM that outputs both an authenticity verdict and descriptive forgery clues. FakeReasoning~\cite{gao2026toward} employs a dual-branch visual encoder with a forgery-aware fusion module to improve visual understanding. AIGI-Holmes~\cite{zhou2025holmes} introduces a three-stage pipeline and a collaborative decoding strategy that integrates perceptual and semantic reasoning for explainable decisions. However, their explanations lack spatial alignment with specific suspicious regions.

Evidence-grounded methods further connect authenticity decisions with localized visual evidence. FakeShield~\cite{xu2025fakeshield} incorporates an external segmentation component to identify suspicious regions, but focuses mainly on manipulated regions rather than fully AI-generated images. LEGION~\cite{kang2025legion} combines an MLLM with a grounding encoder and a pixel decoder to perform detection, artifact segmentation, and explanation generation. FakeXplainer~\cite{ji2026fakexplain} uses supervised fine-tuning and progressive reinforcement learning to jointly generate verdicts, bounding boxes, and explanations; its localization reward is scheduled according to training progress. Locate-Then-Examine~\cite{ji2026lte} first proposes suspicious regions and then re-examines cropped regions, improving access to local details at the cost of multi-stage inference. Although these methods provide localized outputs, they do not explicitly verify whether the authenticity verdict truly depends on the predicted visual evidence.

\subsection{AI-Generated Image Datasets}
\label{sec:supp_datasets}

Existing benchmarks such as GenImage~\cite{zhu2023genimage}, AIGIDataset~\cite{xiao2025AIGIDataset}, Chameleon~\cite{yan2025Chameleon}, and UniversalFakeDetect~\cite{ojha2023towards} mainly support image-level classification and cross-generator evaluation. More recent datasets including FakeClue~\cite{wen2026fakeclue}, MMFR-Dataset~\cite{gao2026toward}, GenExplain~\cite{wu2026explainable}, and FakeBench~\cite{li2025fakebench} add natural-language explanations. However, they lack spatial localization annotations for suspicious regions and therefore do not explicitly align their textual explanations with the corresponding visual areas. This provides insufficient supervision for learning and evaluating whether a model can identify localized visual evidence.

Several datasets provide finer spatial annotations for generated images. PAL4VST~\cite{zhang2023perceptual} and RichHF-18K~\cite{liang2024rich} use masks or points to localize perceptual defects, but they are designed mainly for generation quality assessment and synthesis feedback rather than authenticity detection, and do not provide region-aligned explanations. LOKI~\cite{ye2025loki} and SynthScars~\cite{kang2025legion} provide bounding boxes or masks together with evidence categories or descriptions, yet organize images into single-category scenes, which limits compositional diversity. X-AIGD~\cite{X-AIGD} annotates suspicious regions with evidence categories without providing region-level textual explanations. FakeXplained~\cite{ji2026fakexplain} pairs localized regions with textual explanations, yet its generated images are synthesized from short hand-crafted templates such as ``a photo of [class]'', yielding simple compositions that poorly reflect the richness of real human-centric daily scenes. Overall, these datasets still provide limited coverage of high-resolution images, recent generators, and complex human-centric daily scenes, and seldom offer auditable annotations that jointly align localized visual evidence with evidence categories and region-level explanations.

\section{HAVE Dataset Details}
\label{sec:have_dataset_details}

Following the four-stage pipeline in the main paper (Figure~2), this section supplements the HAVE dataset description in the main paper with implementation details and dataset statistics. Figure~\ref{fig:supp_pipeline} illustrates the overall construction flow. The pipeline is organized as: (1)~Visual Description, (2)~Synthetic Image Generation, (3)~Visual Evidence Annotation, and (4)~Annotation Refinement.

\subsection{Stage 1: Visual Description}
\label{sec:supp_visual_description}

This stage converts high-resolution human-centric real photographs from PortraitCraft into structured visual descriptions. Instead of hand-crafted templates or unconstrained captions, we use GPT-4o to produce objective descriptions covering six semantic aspects. The resulting descriptions have an average length of 147 words and serve as generation inputs for Stage~2.

\begin{figure}[t]
\centering
\includegraphics[width=\linewidth]{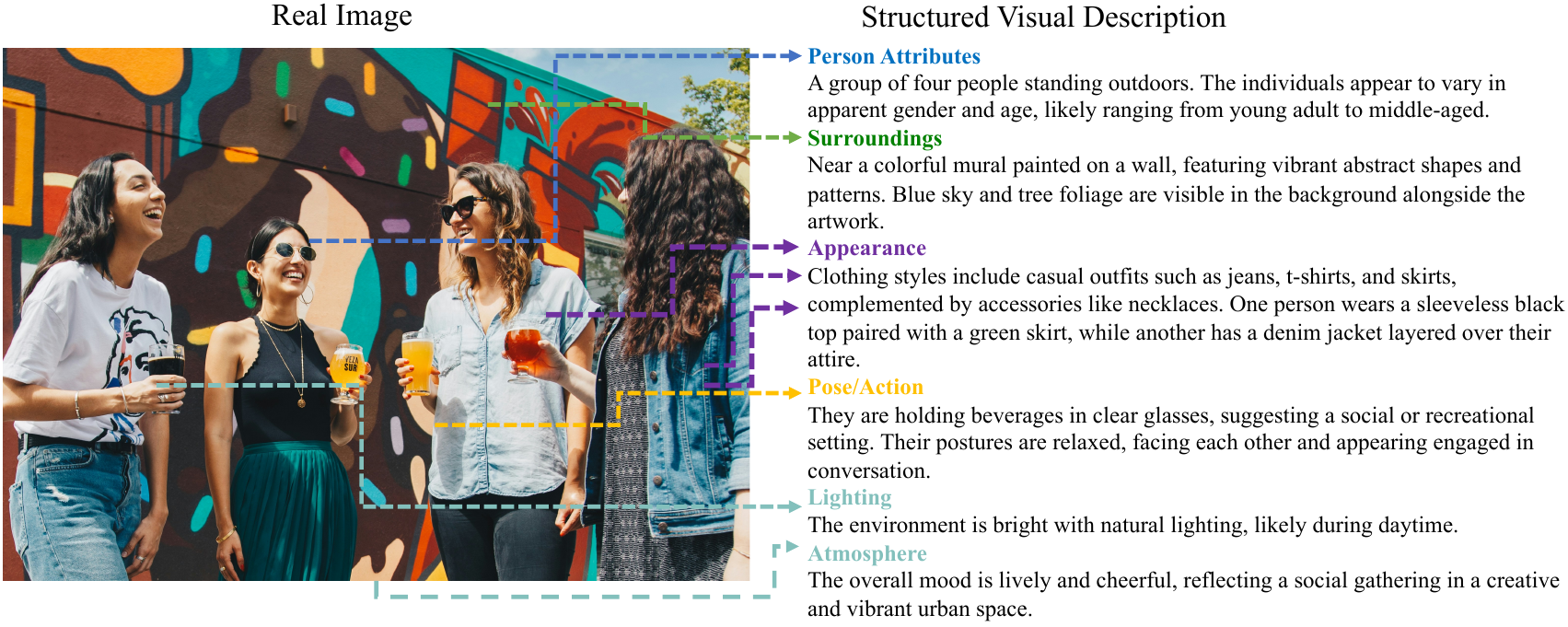}
\caption{Example of a structured visual description in Stage~1. GPT-4o first produces objective descriptions along six semantic aspects, which are then integrated into one fluent paragraph as input for downstream image generation.}
\label{fig:supp_structure_prompt}
\end{figure}

Figure~\ref{fig:supp_structure_prompt} illustrates the structured visual descriptions produced by GPT-4o in Stage~1. Given a human-centric source photograph, the model generates an objective English description organized around six semantic aspects: \textit{person}, \textit{appearance}, \textit{pose/action}, \textit{surroundings}, \textit{lighting}, and \textit{atmosphere}. As shown in Figure~\ref{fig:supp_structure_prompt}, the output captures person-related attributes, clothing and accessories, scene context, illumination conditions, actions and interactions, and overall mood within a single coherent paragraph. This design avoids short template captions and preserves the compositional complexity of real-world human-centric scenes.

\begin{figure}[t]
\centering
\includegraphics[width=\linewidth]{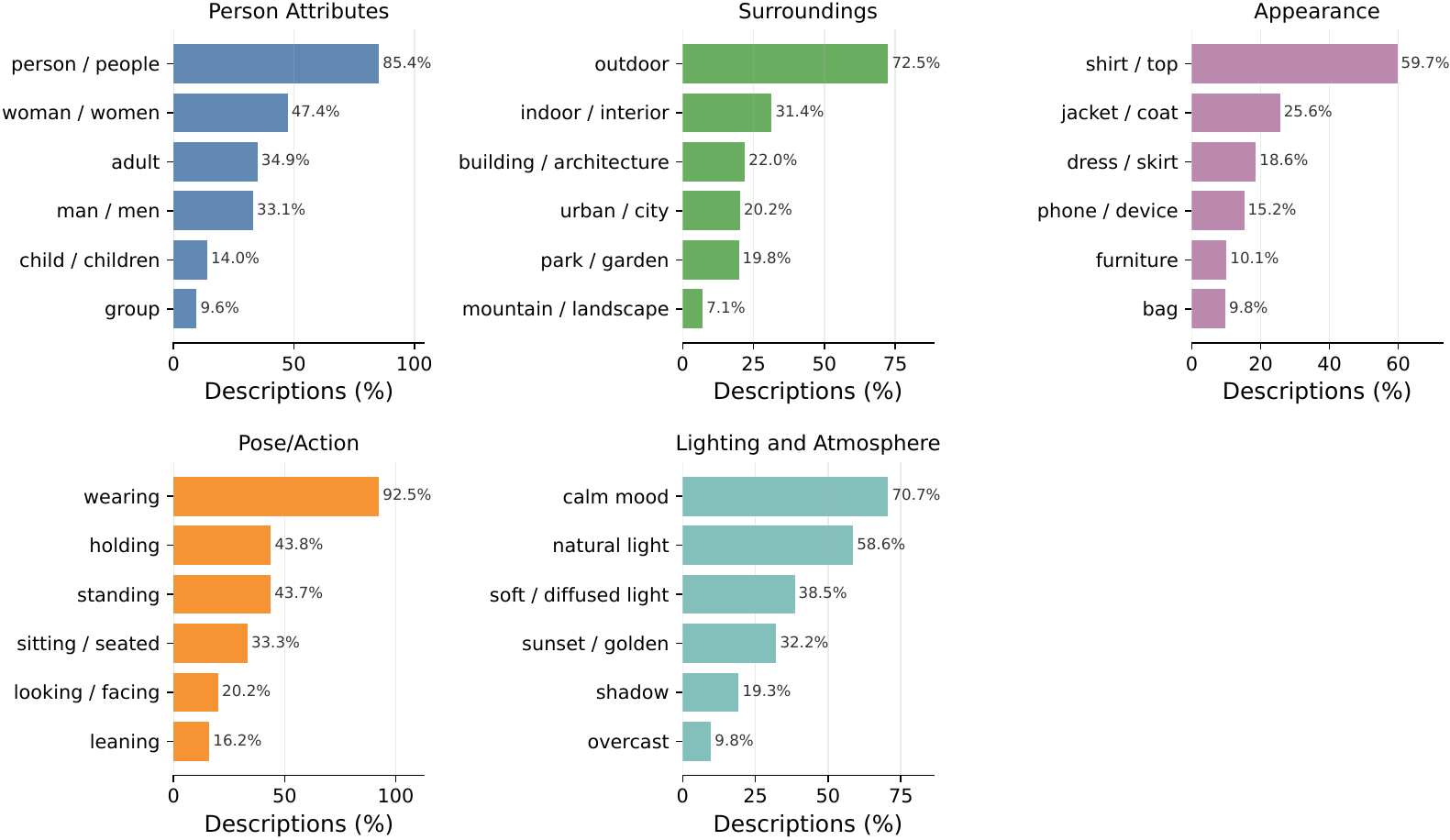}
\caption{Coverage of structured visual descriptions across six semantic dimensions. Each group reports the proportion of descriptions that mention related vocabulary, including person attributes, surroundings, appearance, pose/action, lighting and atmosphere, with representative top terms shown.}
\label{fig:supp_vocab_coverage}
\end{figure}

Figure~\ref{fig:supp_vocab_coverage} further reports, for each semantic group, the proportion of generated structured visual descriptions that mention related vocabulary (person attributes, surroundings, appearance, pose/action, and lighting and atmosphere), indicating that Stage~1 outputs capture diverse human-centric semantics beyond simple class templates.

\subsection{Stage 2: Image Generation}
\label{sec:supp_image_generation}

\paragraph{Controlled generation settings.}
Table~\ref{tab:supp_generators} summarizes the ten generators used in Stage~2 and their controlled generation settings. Each generator receives the Stage~1 structured description as the sole text condition, without additional style suffixes or template wrappers. The selected models span both commercial APIs and open-source systems released in recent years, covering diffusion- and autoregressive-style pipelines with different native aspect-ratio and resolution supports. GPT-Image-1/2 further expose explicit quality levels (low/medium/high), while Seedream-4.0 adopts a fixed 4:3 layout across 1K/2K/4K tiers. This design yields diverse output geometry and rendering characteristics under the same structured prompts, which is important for evaluating detector robustness beyond a single generator family or fixed canvas size. Each structured description is assigned to exactly one generator, so every prompt produces one image without cross-generator sharing.

\begin{table}[t]
\centering
\footnotesize
\setlength{\tabcolsep}{2pt}
\begin{adjustbox}{width=\columnwidth}
\begin{tabular}{l|c|c|c}
\toprule
Generator & Aspect Ratio & Resolution & Quality \\
\midrule
NanoBanana-2  & 1:1, 16:9, 9:16 & 1K, 2K, 4K & - \\
GPT-Image-1   & 1:1, 3:2, 2:3 & 1K, 2K & low, medium, high \\
GPT-Image-2   & 1:1, 3:2, 2:3 & 1K, 2K & low, medium, high \\
NanoBanana-1  & 1:1, 16:9, 9:16 & 1K, 2K, 4K & - \\
Qwen-Z        & 1:1, 16:9, 9:16 & 1K & - \\
Seedream-3.0  & 1:1 & 1K, 2K & - \\
Seedream-4.0  & 4:3 & 1K, 2K, 4K & - \\
Seedream-4.5  & 1:1, 16:9, 9:16 & 1K, 2K, 4K & - \\
Wan-2.6       & 1:1, 3:2, 2:3 & 1K, 2K & - \\
Wan-2.7       & 1:1, 3:2, 2:3 & 1K, 2K & - \\
\bottomrule
\end{tabular}
\end{adjustbox}
\caption{Generators and controlled generation settings in HAVE (Stage~2). Resolution entries denote approximate tiers (1K${\approx}1024$px, 2K${\approx}1536$--$2304$px, 4K${\approx}4096$--$4704$px on the short side). Only GPT-Image-1 and GPT-Image-2 support explicit quality levels.}
\label{tab:supp_generators}
\end{table}

\begin{figure*}[t]
\centering
\includegraphics[width=\textwidth]{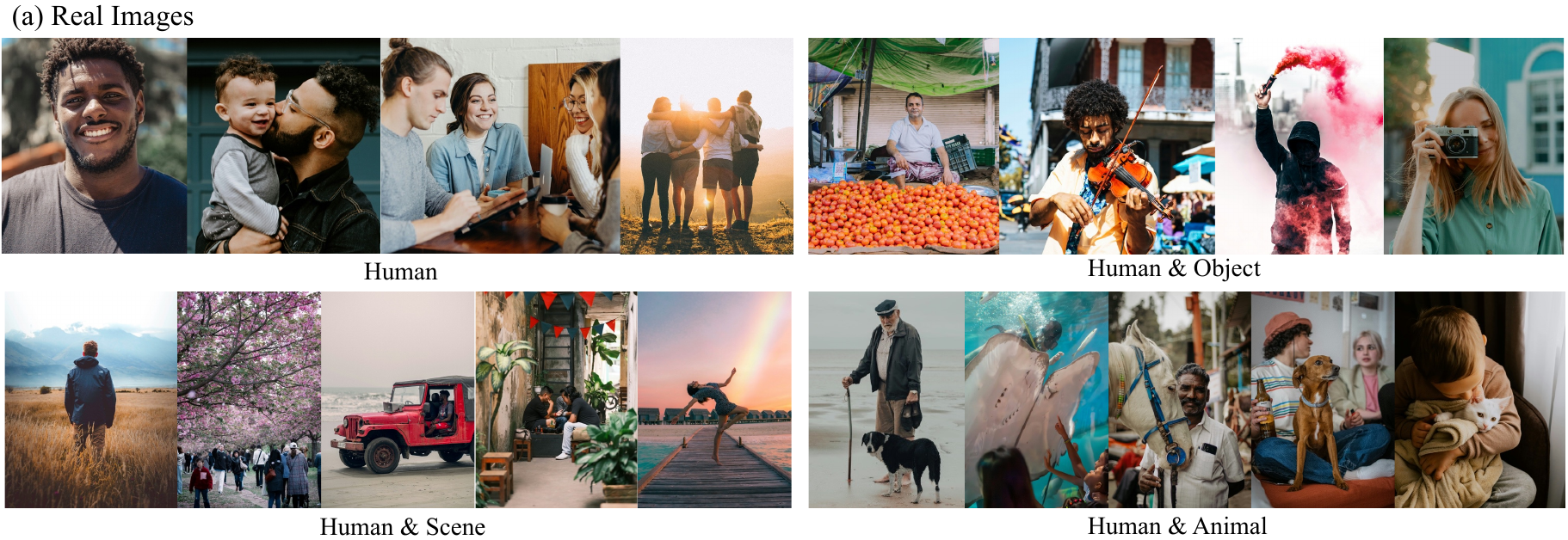}\\[4pt]
\includegraphics[width=\textwidth]{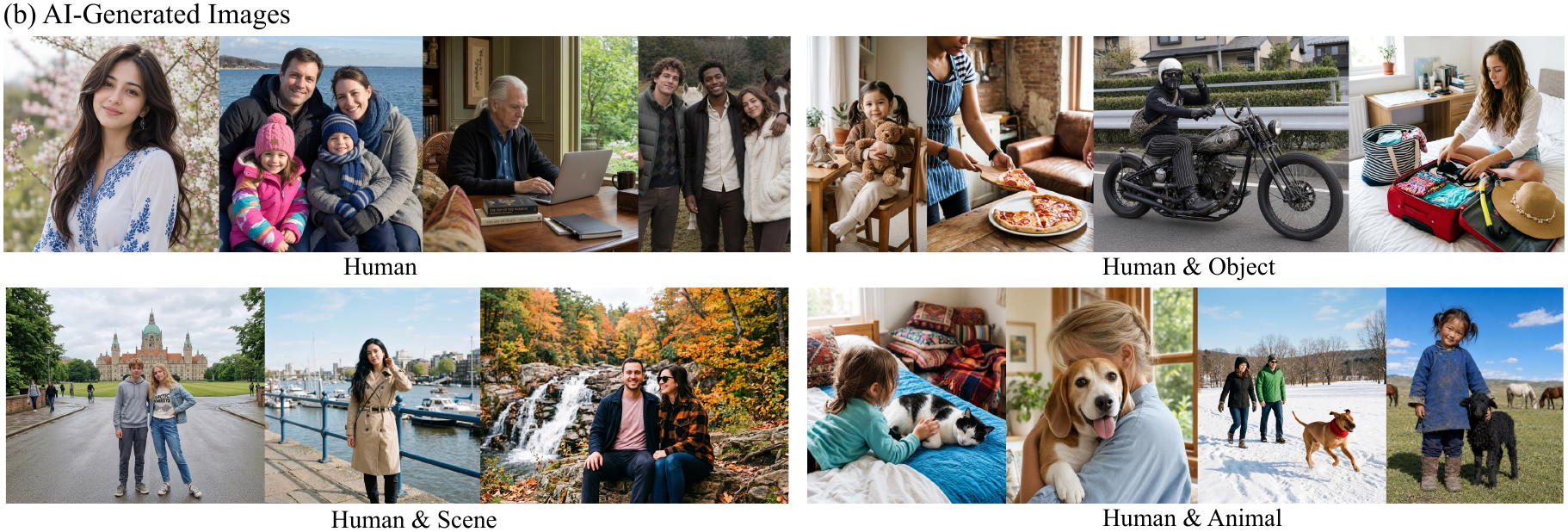}
\caption{Representative real and AI-generated images in HAVE across five human-centric interaction types (\textbf{a} top: real images; \textbf{b} bottom: AI-generated images). Examples cover indoor and outdoor settings, people of diverse skin tones, ages, genders, and group sizes, and interactions with scenery, objects, and animals in complex everyday scenes.}
\label{fig:supp_scene_showcase}
\end{figure*}

\paragraph{Real and generated image showcase.}
Figure~\ref{fig:supp_scene_showcase} illustrates the scene diversity in HAVE across five human-centric interaction types: single-person portraits, multi-person interactions, human--scene compositions, human--animal interactions, and human--object interactions (\textbf{a} top: real images; \textbf{b} bottom: generated images). Real images are collected from PortraitCraft~\cite{sha2026portraitcraft}. Generated images are synthesized in Stage~2 from the corresponding structured visual descriptions using ten recent generators under diverse aspect ratios and resolution settings.

\subsection{Stage 3: Visual Evidence Annotation}
\label{sec:supp_evidence_annotation}

Gemini-3.1-pro inspects each generated image and produces candidate localized visual evidence annotations. Each annotation contains three components: (i)~a normalized bounding box in $[0,1000]$ coordinates with sufficient surrounding context, (ii)~one of eight evidence categories, including Anatomy, Text, Perspective, Interaction, Surface, Lighting, Physical, and Other, and (iii)~a concise textual evidence note describing the visible abnormality inside or around the boxed region.

\begin{figure*}[t]
\centering
\includegraphics[width=\textwidth]{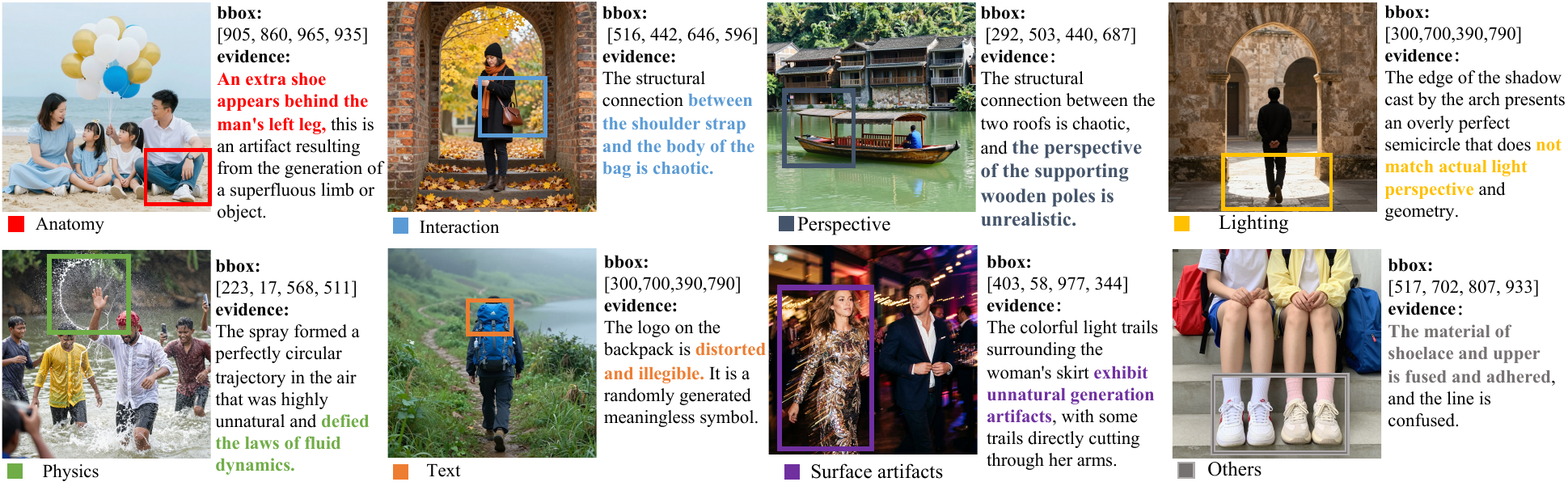}
\caption{Representative visual evidence annotation examples across eight evidence categories in HAVE. Each annotation includes a localized bounding box, a category label, and a region-aligned explanation.}
\label{fig:supp_annotation_example}
\end{figure*}

\paragraph{Annotation examples.}
Figure~\ref{fig:supp_annotation_example} presents eight representative annotated samples covering all eight evidence categories, including Anatomy, Text, Perspective, Interaction, Surface, Lighting, Physical, and Other. Each sample shows a generated human-centric image with localized bounding boxes, category labels, and region-aligned evidence notes, illustrating the format and granularity of evidence annotations.

\subsection{Stage 4: Annotation Refinement}
\label{sec:supp_refinement}

\begin{figure}[t]
\centering
\includegraphics[width=\linewidth]{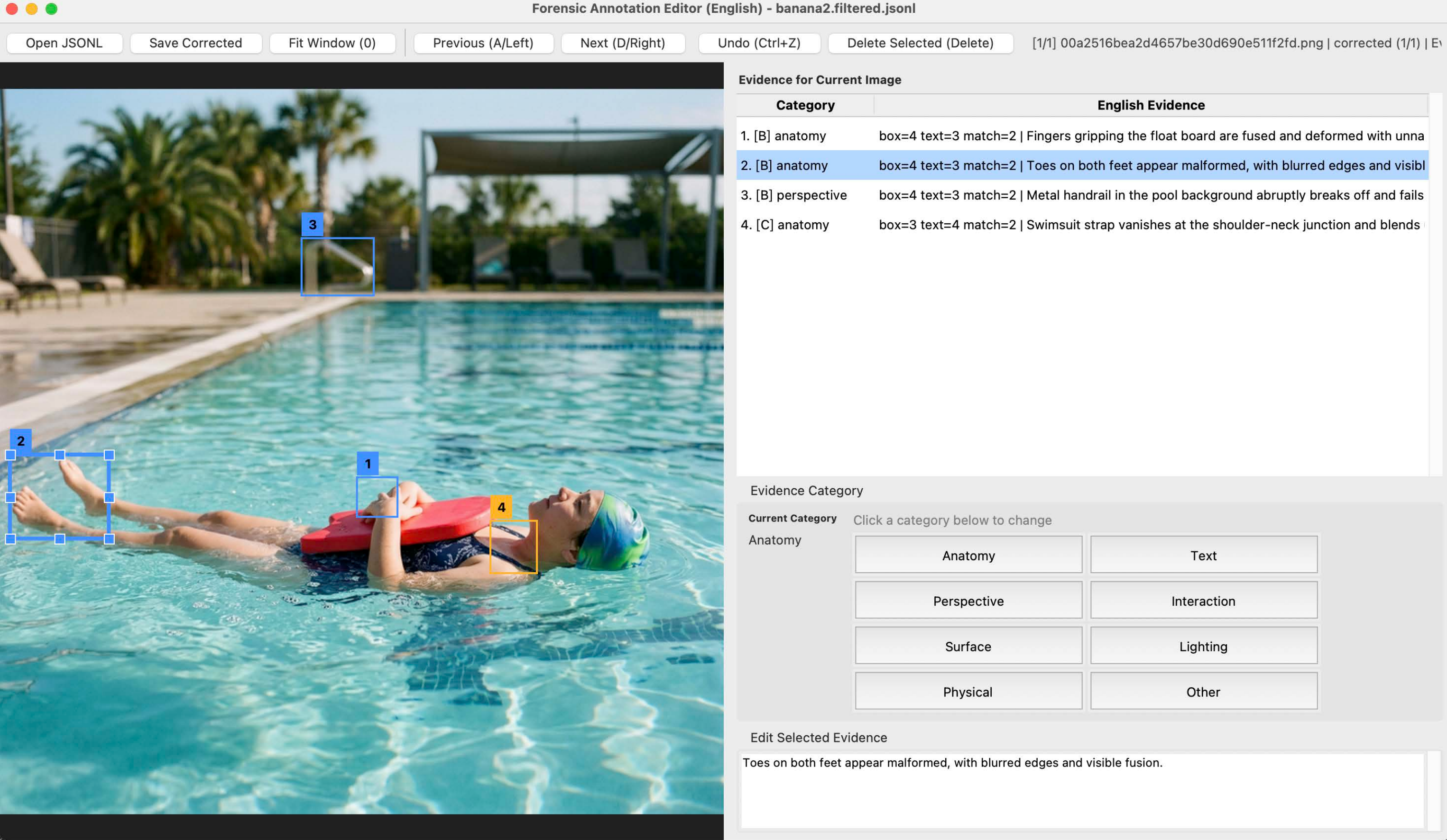}
\caption{Custom annotation refinement interface for HAVE. The tool displays the original image with bounding-box overlay, evidence category and note, and three Qwen3-VL aspect scores for review. Reviewers can manually adjust the bounding box, category, and explanation.}
\label{fig:supp_refinement_ui}
\end{figure}

Candidate annotations from Stage~3 are reviewed through an auditable annotation pipeline, consistent with the main paper. For each candidate annotation, Qwen3-VL~\cite{bai2025qwen3} assigns three aspect scores on a 1-to-5 scale: \textit{box--text match}, \textit{artifact-in-box}, and \textit{description-exists}, measuring whether the box and the evidence refer to the same visual evidence, whether the boxed region contains visible visual evidence, and whether the textual evidence is visually grounded in the image, respectively. Reviewers inspect each candidate through three views: the original image, the image with a red bounding-box overlay, and the cropped in-box region, with the claimed category and evidence note. Figure~\ref{fig:supp_refinement_ui} shows our custom refinement interface used in this process.

Each candidate is then assigned to one mutually exclusive outcome using a priority-based rule with a threshold of 4 for all three scores: \textbf{(A) Matched}: the box--text match score is 4 or 5 $\rightarrow$ keep as a reliable annotation; \textbf{(B) Box-valid}: otherwise, if the artifact-in-box score is 4 or 5 $\rightarrow$ keep, and the evidence note may be manually corrected; \textbf{(C) Re-box}: otherwise, if the description-exists score is 4 or 5 $\rightarrow$ send to Gemini for evidence-guided re-localization, followed by manual inspection; \textbf{Reject}: otherwise, when none of the above conditions is met $\rightarrow$ discard. This priority-based refinement reduces hallucinated evidence, spatial misalignment, and weak boxes while preserving auditable localized visual evidence annotations.

\subsection{Dataset Statistics and Splits}
\label{sec:supp_statistics}

Table~\ref{tab:supp_generator_counts} reports the final per-generator split after Stage~4 refinement. The training set contains 30{,}000 real and 28{,}615 generated images, while the test set contains 10{,}000 real and 10{,}844 generated images, yielding a nearly balanced real/fake distribution for authenticity detection. Real images are evenly allocated at 3{,}000 per generator for training and 1{,}000 per generator for testing, ensuring that each generator subset is paired with the same amount of real-scene supervision. Generated-image counts vary by generator API availability and filtering outcomes---for example, Seedream-4.5 and Seedream-3.0 retain fewer generated samples than NanoBanana-1 or GPT-Image-1---but all ten generators remain represented in both splits. In total, HAVE provides 106{,}609 localized visual evidence instances with bounding boxes, one of eight evidence categories, and region-aligned evidence notes, supporting joint evaluation of detection, grounding, and explanation.

\begin{table}[t]
\centering
\footnotesize
\setlength{\tabcolsep}{3pt}
\begin{adjustbox}{width=\columnwidth}
\begin{tabular}{l|rr|rr|r}
\toprule
\multirow{2}{*}{Generator} & \multicolumn{2}{c|}{Train} & \multicolumn{2}{c|}{Test} & \multirow{2}{*}{\shortstack{Evidence\\instances}} \\
\cmidrule(lr){2-3}\cmidrule(lr){4-5}
 & Real & Gen. & Real & Gen. & \\
\midrule
NanoBanana-2  & 3,000 & 3,935 & 1,000 & 1,047 & 13,075 \\
GPT-Image-1   & 3,000 & 3,385 & 1,000 & 1,122 & 13,254 \\
GPT-Image-2   & 3,000 & 3,681 & 1,000 & 1,039 & 11,219 \\
NanoBanana-1  & 3,000 & 3,560 & 1,000 & 1,135 & 14,411 \\
Qwen-Z        & 3,000 & 3,488 & 1,000 & 1,058 & 11,488 \\
Seedream-3.0  & 3,000 & 1,077 & 1,000 & 1,160 & 7,611 \\
Seedream-4.0  & 3,000 & 2,718 & 1,000 & 1,041 & 8,937 \\
Seedream-4.5  & 3,000 & 831 & 1,000 & 1,048 & 4,614 \\
Wan-2.6       & 3,000 & 2,510 & 1,000 & 1,099 & 9,482 \\
Wan-2.7       & 3,000 & 3,430 & 1,000 & 1,095 & 12,518 \\
\midrule
Total         & 30,000 & 28,615 & 10,000 & 10,844 & 106,609 \\
\bottomrule
\end{tabular}
\end{adjustbox}
\caption{Final dataset split in HAVE after annotation refinement. Training real images are evenly allocated at 3{,}000 per generator. Test real images are randomly sampled at 1{,}000 per generator. Generated-image and evidence counts reflect the merged training set and test set.}
\label{tab:supp_generator_counts}
\end{table}

\section{Extended Experimental Results}
\label{sec:supp_experiments}

Due to space limits in the main paper, we provide the full experimental tables and extended result analyses here. All in-domain results are evaluated on HAVE unless otherwise stated. The best and second-best results in each row or column are highlighted in \textbf{bold} and \underline{underlined}, respectively.

\subsection{Training and Evaluation Protocol}
\label{sec:supp_eval_protocol}

Following the main paper, we re-train CNNSpot~\cite{wang2020cnnspot}, UnivFD~\cite{ojha2023towards}, NPR~\cite{tan2024npr}, RINE~\cite{koutlis2024RINE}, AIDE~\cite{yan2025Chameleon}, Effort~\cite{yan2025effort}, DGS-Net~\cite{yan2026dgsnet}, and FakeReasoning~\cite{gao2026toward} on the HAVE training set; adopt official weights for FakeVLM~\cite{wen2026fakeclue}, FakeShield~\cite{xu2025fakeshield}, LEGION~\cite{kang2025legion}, SparseViT~\cite{su2025sparsevit}, and FRD-Net~\cite{chen2026frd}; and include InternVL2.5-26B~\cite{chen2024internvl}, Qwen3-VL-32B~\cite{bai2025qwen3}, and LLaMA3.2-Vision-11B~\cite{grattafiori2024llama}.

For PAVE, we keep the VLM preprocessing and generation configuration aligned between training and testing. Images are processed with Qwen3-VL's dynamic-resolution policy using \texttt{min\_pixels=3136} and \texttt{max\_pixels=1605632}. The maximum input context length is set to \texttt{max\_length=8192}, and the model generates structured outputs with \texttt{max\_completion\_length=1024}. These settings are applied consistently across the SFT and RL stages as well as during evaluation on HAVE and the cross-dataset benchmarks. PAVE also uses the same task instruction throughout SFT, RL, and testing.
For other VLM-based baselines, including FakeVLM, FakeShield, LEGION, and FakeReasoning, we follow the inference settings supported by their official code and released evaluation pipelines. For InternVL2.5-26B, LLaMA3.2-Vision-11B, and Qwen3-VL-32B, we adopt the same evaluation prompt as PAVE at test time. The prompt is:
\noindent\includegraphics[width=\linewidth]{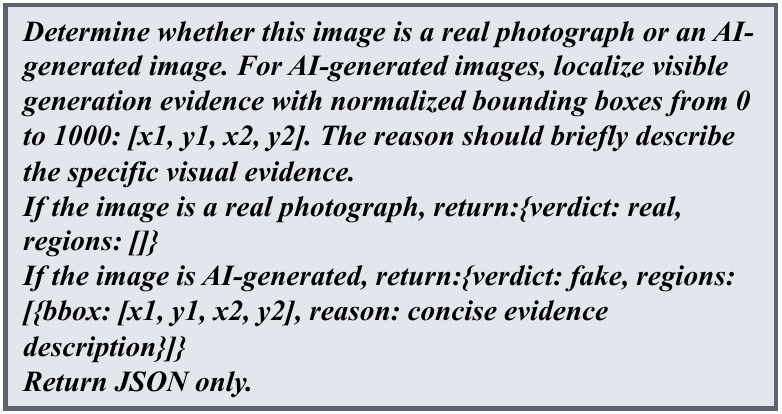}

\subsection{HAVE Results}
\label{sec:supp_have_results}

We report complete in-domain results on the HAVE test set across authenticity detection, visual evidence grounding, and explanation quality. These analyses complement the summary in the main paper and provide detailed results for all compared methods on each of the ten generator subsets in HAVE.

\subsubsection{Authenticity Detection on HAVE}
Table~\ref{tab:supp_detection_acc_transposed} reports per-generator authenticity detection accuracy on HAVE. PAVE achieves the highest average accuracy and ranks first on all ten generator subsets, consistently outperforming both retrained non-LLM detectors and LLM-based baselines. Among the retrained non-LLM methods, DGS-Net performs best overall, followed by AIDE and RINE, while NPR shows larger performance fluctuations across generators, indicating sensitivity to generator-specific rendering characteristics. FakeReasoning also remains clearly behind PAVE despite being retrained on HAVE. PAVE further maintains strong performance on challenging subsets such as SeeDream-4.5, suggesting robust generalization across diverse recent generators. Overall, these results indicate that unified region-aligned training strengthens image-level authenticity classification in addition to improving grounding and explanation quality.

\begin{table*}[!t]
\centering
{\footnotesize
\tabcolsep=6 pt
\begin{adjustbox}{width=\textwidth}
\begin{tabular}{lccccccccccc}
\toprule
Method & Banana2 & GPT-1 & GPT-2 & Banana & Qwen-Z & SeeD-3.0 & SeeD-4.0 & SeeD-4.5 & Wan-2.6 & Wan-2.7 & Avg. \\
\midrule
\multicolumn{12}{l}{\textbf{Non-LLM Based AI Image Detector}} \\
\cmidrule(lr){1-12}
CNNSpot       & 93.32 & 92.27 & 92.05 & 91.14 & 92.09 & 93.09 & 91.68 & 84.32 & 88.49 & 91.82 & 91.03 \\
UnivFD        & 88.77 & 90.77 & 91.73 & 88.27 & 90.27 & 94.00 & 93.59 & 78.36 & 80.76 & 85.27 & 88.18 \\
NPR           & 95.59 & 83.73 & 63.64 & \textbf{96.86} & 95.05 & 83.27 & 70.86 & 62.45 & \textbf{97.50} & 94.14 & 84.31 \\
RINE          & 94.27 & 93.45 & 92.36 & 92.36 & 93.50 & \underline{95.45} & \underline{94.77} & 86.91 & 88.44 & 90.27 & 92.18 \\
AIDE          & 94.60 & 93.60 & 93.40 & 94.10 & 94.70 & 93.60 & 91.60 & 85.80 & 90.10 & 91.40 & 92.30 \\
Effort        & 86.09 & 84.95 & 85.86 & 85.14 & 84.82 & 84.45 & 83.55 & 86.73 & 84.99 & 84.64 & 85.12 \\
DGS-Net       & \underline{95.80} & \underline{95.40} & \underline{93.50} & 95.90 & \textbf{96.10} & 93.50 & 92.00 & \underline{87.50} & 95.80 & \textbf{95.00} & \underline{94.05} \\
\cmidrule(lr){1-12}
\multicolumn{12}{l}{\textbf{LLM Based AI Image Detector}} \\
\cmidrule(lr){1-12}
FakeVLM       & 74.59 & 71.27 & 66.95 & 82.73 & 71.86 & 77.23 & 58.86 & 56.14 & 76.57 & 75.68 & 71.19 \\
FakeReas.     & 68.45 & 70.50 & 72.95 & 67.86 & 65.18 & 76.86 & 64.41 & 57.09 & 64.15 & 68.64 & 67.56 \\
FakeShield    & 65.91 & 65.45 & 68.05 & 66.00 & 64.91 & 85.32 & 46.82 & 65.64 & 66.88 & 66.14 & 66.11 \\
\midrule
\textbf{PAVE(Ours)} & \textbf{98.73} & \textbf{96.51} & \textbf{96.56} & \underline{96.06} & \underline{95.62} & \textbf{96.94} & \textbf{94.90} & \textbf{94.92} & \underline{96.09} & \underline{94.79} & \textbf{96.11} \\
\bottomrule
\end{tabular}
\end{adjustbox}
\captionof{table}{Authenticity detection performance (Acc, \%) of different methods on HAVE (methods as rows).}
\label{tab:supp_detection_acc_transposed}

\vspace{4pt}
\tabcolsep=2 pt
\begin{adjustbox}{width=\linewidth}
\begin{tabular}{l|*{10}{cc|}cc}
\toprule
\multirow{2}{*}{Method} & \multicolumn{2}{c|}{Banana2} & \multicolumn{2}{c|}{GPT-1} & \multicolumn{2}{c|}{GPT-2} & \multicolumn{2}{c|}{Banana} & \multicolumn{2}{c|}{Qwen-Z} & \multicolumn{2}{c|}{SeeD-3.0} & \multicolumn{2}{c|}{SeeD-4.0} & \multicolumn{2}{c|}{SeeD-4.5} & \multicolumn{2}{c|}{Wan-2.6} & \multicolumn{2}{c|}{Wan-2.7} & \multicolumn{2}{c}{Avg.} \\
                        & IoU & F1 & IoU & F1 & IoU & F1 & IoU & F1 & IoU & F1 & IoU & F1 & IoU & F1 & IoU & F1 & IoU & F1 & IoU & F1 & IoU & F1 \\
\midrule
InternVL2.5      & 2.18 & 4.04 & 3.53 & 6.07 & 1.83 & 3.33 & 3.33 & 5.79 & 3.54 & 5.79 & 1.44 & 2.61 & 1.95 & 3.63 & 2.70 & 4.54 & 1.75 & 3.17 & 3.61 & 5.53 & 2.59 & 4.45 \\
Qwen3-VL         & \underline{10.07} & \underline{16.67} & \underline{17.80} & \underline{26.65} & \underline{10.29} & \underline{16.69} & \underline{12.88} & \underline{20.64} & \underline{12.28} & \underline{19.30} & \underline{12.92} & \underline{20.65} & \underline{12.37} & \underline{19.95} & \underline{13.94} & \underline{22.21} & \underline{14.75} & \underline{23.17} & \underline{12.51} & \underline{20.13} & \underline{12.98} & \underline{20.61} \\
LLaMA3.2-Vision  & 3.57 & 6.65 & 7.16 & 12.61 & 4.24 & 7.76 & 6.50 & 11.56 & 4.92 & 8.67 & 5.42 & 9.76 & 5.28 & 9.56 & 6.08 & 10.99 & 6.51 & 11.59 & 7.12 & 12.14 & 5.68 & 10.13 \\
FRD-Net          & 5.43 & 9.65 & 9.12 & 15.60 & 6.15 & 10.71 & 7.30 & 12.67 & 6.29 & 11.01 & 7.61 & 13.41 & 6.85 & 11.94 & 7.76 & 13.33 & 8.23 & 14.09 & 7.61 & 13.13 & 7.23 & 12.55 \\
SparseViT        & 5.10 & 9.20 & 9.71 & 16.51 & 5.76 & 10.27 & 6.73 & 11.74 & 5.67 & 10.06 & 7.14 & 12.66 & 6.43 & 11.26 & 7.48 & 12.93 & 7.55 & 13.06 & 7.14 & 12.46 & 6.87 & 12.02 \\
FakeShield       & 4.04 & 7.01 & 6.75 & 11.48 & 4.61 & 7.80 & 5.09 & 8.62 & 3.89 & 6.84 & 5.50 & 9.49 & 4.97 & 8.52 & 4.98 & 8.57 & 5.80 & 9.95 & 5.02 & 8.59 & 5.07 & 8.69 \\
LEGION           & 2.00 & 3.40 & 4.39 & 7.13 & 2.97 & 5.01 & 3.18 & 5.27 & 2.55 & 4.14 & 4.24 & 7.01 & 2.63 & 4.30 & 3.40 & 5.52 & 4.38 & 7.01 & 3.65 & 5.93 & 3.34 & 5.47 \\
\midrule
\textbf{PAVE(Ours)} & \textbf{27.06} & \textbf{38.27} & \textbf{38.23} & \textbf{50.71} & \textbf{26.03} & \textbf{36.74} & \textbf{33.52} & \textbf{45.41} & \textbf{30.47} & \textbf{42.02} & \textbf{37.53} & \textbf{50.29} & \textbf{26.88} & \textbf{37.86} & \textbf{27.48} & \textbf{38.43} & \textbf{33.16} & \textbf{44.75} & \textbf{30.73} & \textbf{42.13} & \textbf{31.11} & \textbf{42.66} \\
\bottomrule
\end{tabular}
\end{adjustbox}
\captionof{table}{Visual evidence grounding performance (IoU and F1, \%) on HAVE (methods as rows).}
\label{tab:supp_localization_ours_v_transposed}

\vspace{4pt}
\tabcolsep=2 pt
\begin{adjustbox}{width=\linewidth}
\begin{tabular}{l|*{10}{cc|}cc}
\toprule
\multirow{2}{*}{Method} & \multicolumn{2}{c|}{Banana2} & \multicolumn{2}{c|}{GPT-1} & \multicolumn{2}{c|}{GPT-2} & \multicolumn{2}{c|}{Banana} & \multicolumn{2}{c|}{Qwen-Z} & \multicolumn{2}{c|}{SeeD-3.0} & \multicolumn{2}{c|}{SeeD-4.0} & \multicolumn{2}{c|}{SeeD-4.5} & \multicolumn{2}{c|}{Wan-2.6} & \multicolumn{2}{c|}{Wan-2.7} & \multicolumn{2}{c}{Avg.} \\
                        & R-L & CSS & R-L & CSS & R-L & CSS & R-L & CSS & R-L & CSS & R-L & CSS & R-L & CSS & R-L & CSS & R-L & CSS & R-L & CSS & R-L & CSS \\
\midrule
InternVL2.5      & 9.51 & 28.98 & 6.54 & 25.69 & 8.12 & 26.29 & 8.72 & 29.63 & 8.20 & 30.21 & 7.86 & 28.82 & 7.65 & 30.64 & 8.49 & 27.44 & 8.20 & 29.44 & 7.10 & 30.98 & 7.98 & 28.70 \\
Qwen3-VL         & \underline{14.96} & \underline{48.15} & \underline{14.86} & \underline{50.32} & \underline{13.29} & \underline{44.83} & \underline{14.93} & \underline{49.34} & \underline{14.69} & \underline{48.97} & \underline{15.34} & \underline{50.23} & \underline{13.30} & \underline{45.19} & \underline{14.74} & \underline{46.24} & \underline{14.68} & \underline{47.12} & \underline{14.96} & \underline{49.54} & \underline{14.85} & \underline{48.99} \\
LLaMA3.2-Vision  & 7.15 & 19.90 & 8.31 & 21.35 & 7.97 & 21.93 & 8.51 & 22.36 & 7.85 & 23.32 & 8.67 & 20.36 & 8.20 & 24.14 & 7.29 & 20.43 & 8.48 & 22.50 & 7.53 & 23.50 & 8.29 & 21.53 \\
FakeReas.        & 5.65 & 23.59 & 5.11 & 25.46 & 5.34 & 24.17 & 5.58 & 24.44 & 5.55 & 24.04 & 4.82 & 26.07 & 5.57 & 24.16 & 5.75 & 23.98 & 5.49 & 24.57 & 5.43 & 24.24 & 5.42 & 24.49 \\
FakeVLM          & 3.80 & 29.73 & 3.74 & 28.64 & 3.79 & 27.59 & 5.16 & 31.58 & 4.23 & 30.52 & 3.16 & 29.04 & 4.15 & 28.01 & 4.70 & 28.03 & 3.79 & 29.93 & 4.04 & 29.15 & 4.05 & 29.24 \\
FakeShield       & 4.80 & 31.18 & 5.38 & 33.05 & 4.38 & 29.72 & 5.19 & 32.17 & 4.53 & 31.63 & 5.62 & 33.14 & 4.27 & 30.78 & 4.47 & 31.49 & 4.99 & 31.80 & 4.88 & 31.12 & 4.97 & 31.80 \\
LEGION           & 7.58 & 44.26 & 8.87 & 47.85 & 6.71 & 41.76 & 8.57 & 46.51 & 7.59 & 45.57 & 9.36 & 48.66 & 7.03 & 43.34 & 7.09 & 43.60 & 7.69 & 45.14 & 7.66 & 44.64 & 7.81 & 45.13 \\
\textbf{PAVE(Ours)} & \textbf{21.08} & \textbf{61.51} & \textbf{22.68} & \textbf{65.95} & \textbf{19.78} & \textbf{57.83} & \textbf{22.27} & \textbf{65.34} & \textbf{21.02} & \textbf{62.24} & \textbf{23.99} & \textbf{68.20} & \textbf{20.22} & \textbf{59.58} & \textbf{20.18} & \textbf{59.66} & \textbf{21.62} & \textbf{63.03} & \textbf{21.26} & \textbf{62.93} & \textbf{21.45} & \textbf{62.74} \\
\bottomrule
\end{tabular}
\end{adjustbox}
\captionof{table}{Explanation performance (ROUGE-L abbreviated as R-L and CSS, \%) on HAVE (methods as rows).}
\label{tab:supp_explanation_ours_v_transposed}
}
\end{table*}

\subsubsection{Visual Evidence Grounding on HAVE}
Table~\ref{tab:supp_localization_ours_v_transposed} reports visual evidence grounding performance in terms of IoU and F1. PAVE achieves an average IoU of 31.11\% and F1 of 42.66\%, more than doubling the strongest baseline Qwen3-VL. The improvement is consistent across all ten generator subsets, with particularly large gains on subsets such as GPT-Image-1 and SeeDream-3.0. General-purpose VLMs and forgery-specialized models, including InternVL2.5, LLaMA3.2-Vision, FakeShield, and LEGION, remain near or below 10\% F1 on average, indicating that a correct authenticity label does not necessarily translate into accurate evidence localization in complex human-centric scenes. FRD-Net and SparseViT perform slightly better than these baselines but still fall far short of PAVE. Overall, these results validate the difficulty of grounding fine-grained visual evidence and demonstrate the benefit of region-aligned supervision on HAVE.

\subsubsection{Explanation Quality on HAVE}
Table~\ref{tab:supp_explanation_ours_v_transposed} reports explanation quality using ROUGE-L and CSS. PAVE achieves the best average CSS of 62.74\% and ROUGE-L of 21.45\%, outperforming Qwen3-VL by 13.75 and 6.60 percentage points, respectively. Qwen3-VL is the second-best method on both metrics, followed by LEGION. FakeVLM and FakeReasoning obtain moderate CSS but much lower ROUGE-L, suggesting that their generated text often overlaps weakly with ground-truth evidence wording despite partially relevant semantics. PAVE also leads on every generator subset, with particularly strong explanation quality on SeeDream-3.0, where localized artifacts are visually salient. Overall, these results indicate that PAVE produces explanations that are both semantically consistent with ground-truth evidence and lexically aligned with reference descriptions.

\subsection{Cross-Dataset Generalization}
\label{sec:supp_cross_dataset}

To assess transferability beyond the training generators and human-centric scenes in HAVE, we evaluate PAVE and VLM-based baselines on four out-of-distribution datasets with spatial annotations: SynthScars~\cite{kang2025legion}, LOKI~\cite{ye2025loki}, X-AIGD~\cite{X-AIGD}, and RichHF~\cite{liang2024rich}. The first three datasets provide paired artifact regions and textual explanations, whereas RichHF contains bounding-box annotations for perceptual defects but no region--explanation pairs and is therefore included only in the grounding evaluation. All models are evaluated zero-shot without fine-tuning on these benchmarks.

\subsubsection{Visual Evidence Grounding}
Table~\ref{tab:supp_localization} reports cross-dataset visual evidence grounding. PAVE achieves the best overall performance with an average IoU of 19.04\% and F1 of 29.12\%, outperforming the second-best method by +5.80\% IoU and +8.64\% F1. It obtains the highest F1 on all four datasets and the highest IoU on RichHF, LOKI, and X-AIGD. On SynthScars, LEGION achieves a slightly higher IoU, which is expected because SynthScars is an in-distribution benchmark for LEGION; nevertheless, PAVE still yields the best F1, indicating more reliable box--evidence matching under the joint evaluation protocol. Qwen3-VL remains a strong general-purpose baseline, whereas FakeShield shows limited transfer. The largest gains over Qwen3-VL appear on RichHF (+7.53\% IoU / +11.42\% F1) and LOKI (+6.28\% IoU / +10.66\% F1), where artifact categories and scene compositions differ substantially from HAVE. These results suggest that region-aligned training on complex human-centric scenes helps the model localize visible evidence even when generator families, resolutions, and annotation styles change.

\begin{table*}[!t]
  \centering
  \begin{minipage}[t]{0.615\textwidth}
  \centering
  \tabcolsep=3 pt
  \begin{adjustbox}{width=\linewidth}
  \begin{tabular}{l|rr|rr|rr|rr|rr}
  \toprule
  \multirow{2}{*}{Method}
    & \multicolumn{2}{c|}{SynthScars}
    & \multicolumn{2}{c|}{RichHF}
    & \multicolumn{2}{c|}{LOKI}
    & \multicolumn{2}{c|}{X-AIGD}
    & \multicolumn{2}{c}{Avg.} \\
  & IoU & F1 & IoU & F1 & IoU & F1 & IoU & F1 & IoU & F1 \\
  \midrule
  InternVL2.5 & 1.47 & 2.33 & 3.19 & 5.73 & 3.23 & 5.23 & 10.67 & 2.85 & 4.64 & 4.04 \\
  LLaMA3.2-Vision & 3.16 & 5.09 & 3.81 & 6.85 & 9.31 & 14.26 & 12.20 & 7.51 & 7.12 & 8.43 \\
  Qwen3-VL & 9.89 & 14.76 & \underline{11.59} & \underline{19.28} & 11.65 & 16.29 & \underline{15.12} & \underline{15.86} & 12.06 & 16.55 \\
  FakeShield & 7.66 & 11.54 & 6.07 & 10.78 & \underline{13.81} & \underline{21.55} & 6.23 & 9.49 & 8.44 & 13.34 \\
  LEGION & \textbf{22.35} & \underline{32.23} & 11.41 & 19.23 & 9.87 & 15.99 & 9.32 & 14.47 & \underline{13.24} & \underline{20.48} \\
  \midrule
  \textbf{PAVE(Ours)} & \underline{22.20} & \textbf{33.20} & \textbf{19.12} & \textbf{30.70} & \textbf{17.93} & \textbf{26.95} & \textbf{16.91} & \textbf{25.64} & \textbf{19.04} & \textbf{29.12} \\
  \bottomrule
  \end{tabular}
  \end{adjustbox}
  \caption{Visual evidence grounding performance (IoU and F1, \%) across cross-dataset benchmarks.}
  \label{tab:supp_localization}
  \end{minipage}\hfill
  \begin{minipage}[t]{0.365\textwidth}
  \centering
  \tabcolsep=4 pt
  \begin{adjustbox}{width=\linewidth}
  \begin{tabular}{l|c|c|c|c|c}
  \toprule
  Method & Params & SynthScars & LOKI & X-AIGD & Avg. \\
  \midrule
  InternVL2.5 & 26B & 9.30 & 14.02 & 2.96 & 8.76 \\
  LLaMA3.2-V & 11B & 11.17 & 16.21 & 6.58 & 11.32 \\
  Qwen3-VL & 32B & 26.39 & 21.14 & 10.27 & 19.27 \\
  FakeVLM    & 7B  & 31.53 & \textbf{49.38} & 12.93 & 20.85 \\
  FakeReas.  & 13B & 34.90 & 40.61 & \textbf{14.62} & 22.33 \\
  FakeShield & 22B & 35.44 & 37.79 & 7.88  & 27.04 \\
  LEGION     & 8B  & \textbf{53.49} & \underline{43.48} & \underline{11.74} & \textbf{36.24} \\
  \midrule
  \textbf{PAVE(Ours)} & 4B & \underline{48.42} & \underline{44.19} & \underline{14.47} & \underline{35.69} \\
  \bottomrule
  \end{tabular}
  \end{adjustbox}
  \caption{Explanation performance (CSS, \%) across cross-dataset benchmarks with model parameters.}
  \label{tab:supp_explanation_method}
  \end{minipage}
  \end{table*}

\subsubsection{Explanation Quality}
Table~\ref{tab:supp_explanation_method} reports cross-dataset explanation quality using CSS, together with model size. Although FakeVLM reaches the highest LOKI CSS and FakeReasoning achieves the best X-AIGD CSS, both methods have lower overall averages due to weak performance on other datasets. LEGION attains the best average CSS of 36.24\% and the highest SynthScars score, again benefiting from overlap with that benchmark. PAVE ranks first on LOKI and X-AIGD, second on SynthScars, and achieves a competitive overall CSS of 35.69\% with only 4B parameters, ranking second overall and clearly ahead of Qwen3-VL. This parameter-efficient performance indicates that explicit region--explanation alignment during training helps produce transferable evidence descriptions under domain shift, rather than dataset-specific wording patterns. Notably, PAVE improves LOKI CSS by +23.05 over Qwen3-VL despite using eight times fewer parameters, showing that alignment supervision is more important than model scale for producing grounded explanations on unseen benchmarks.

\subsection{Qualitative Visualization Analysis}
\label{sec:supp_visualization}
\begin{figure*}[!b]
  \centering
  \includegraphics[width=\textwidth]{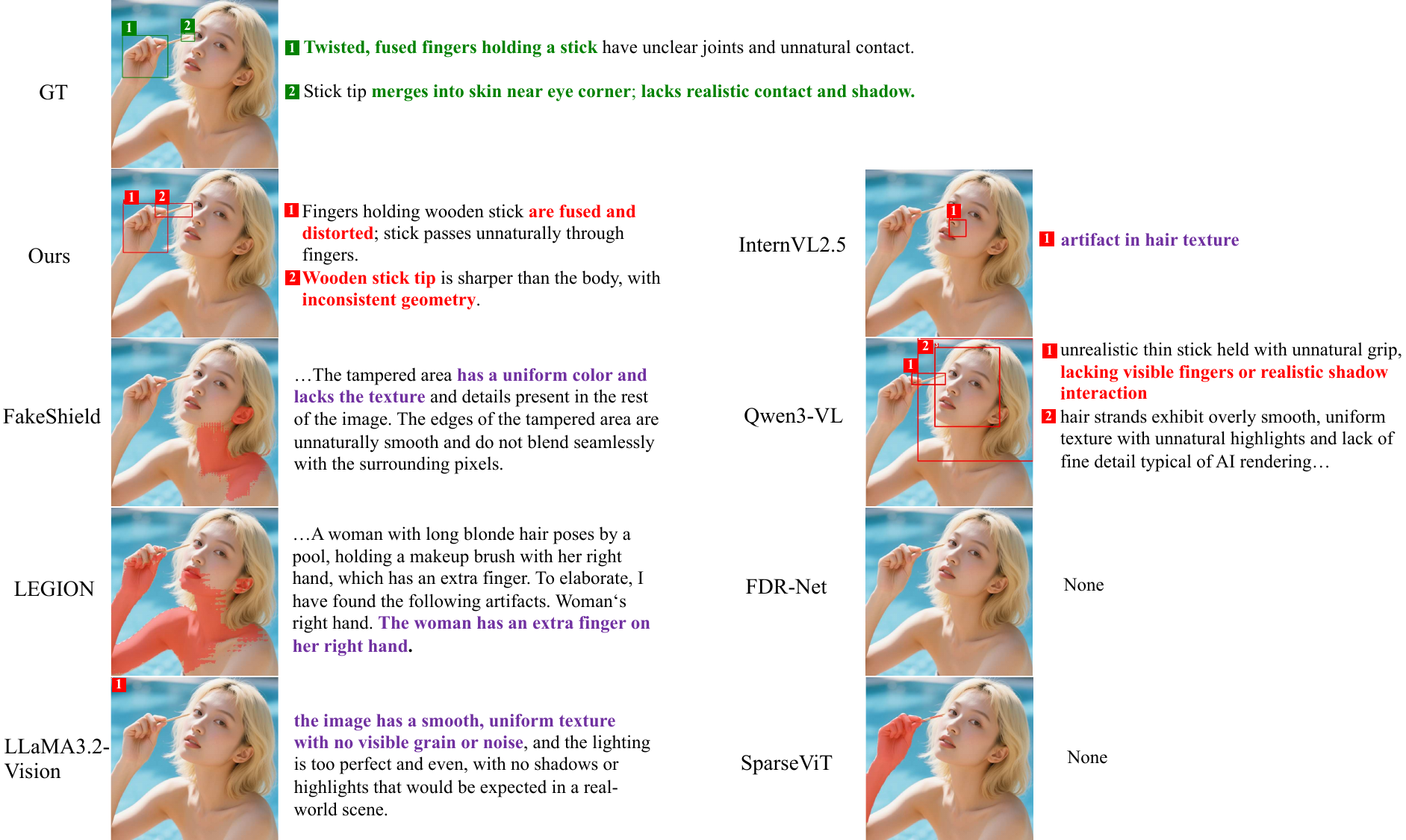}\\[4pt]
  \includegraphics[width=\textwidth]{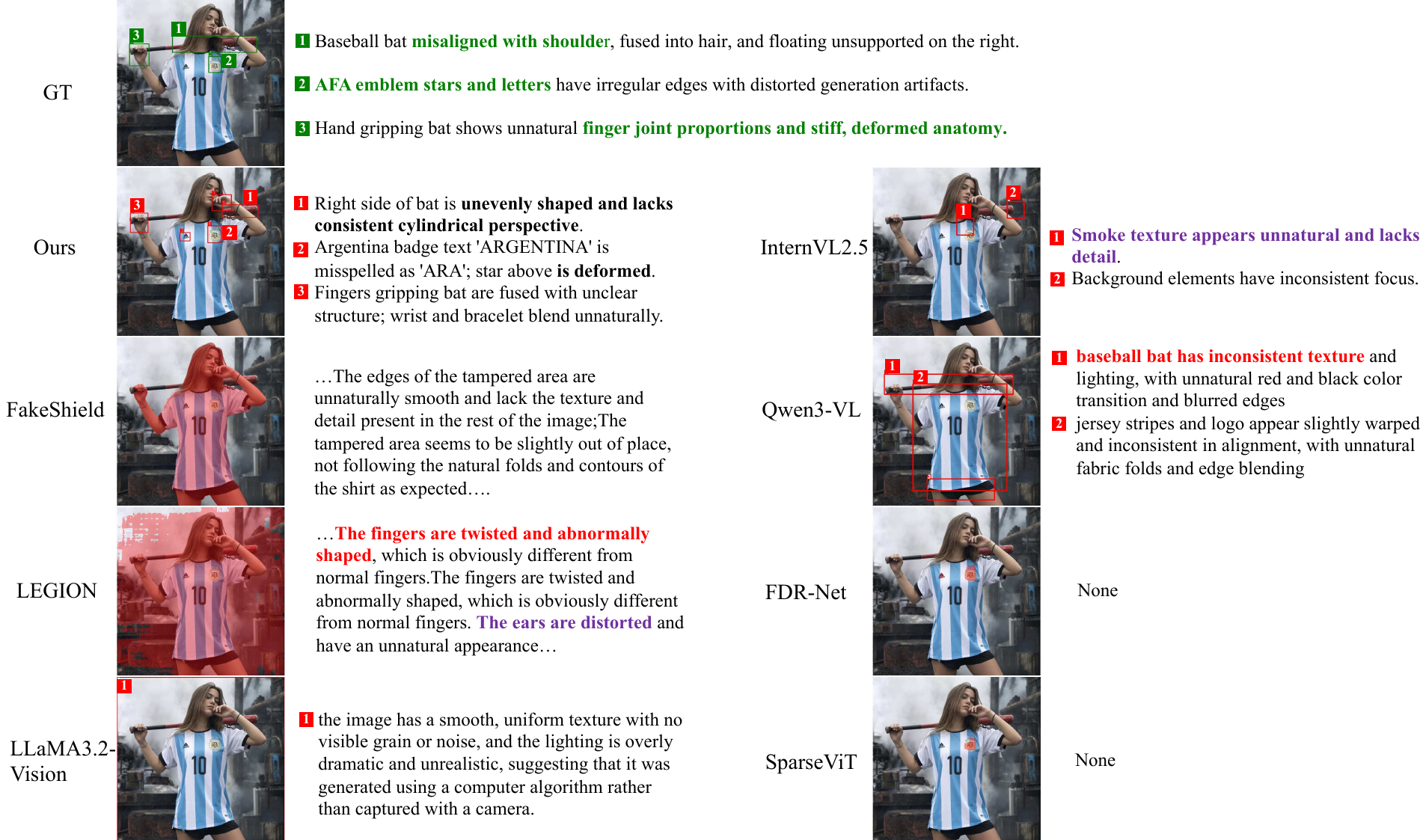}
  \caption{Qualitative comparison of visual evidence grounding and explanation across cross-dataset benchmarks. Each numbered box is paired with its corresponding explanation. Explanations consistent with the localized visual evidence are highlighted in red, whereas mismatched or unreliable explanations are shown in purple.}
  \label{fig:supp_visual_cross_methods}
  \end{figure*}
We provide extended qualitative analyses to complement the quantitative results above and the in-domain visualization presented in the main paper. For each example, we display the predicted evidence boxes together with their paired explanations. Following the main paper, text highlighted in \textbf{red} indicates that the explanation is visually grounded in the corresponding boxed region, whereas text in \textbf{purple} denotes an unreliable explanation that does not match the localized content.

The main-paper visualization compares PAVE with LEGION and Qwen3-VL-32B on representative HAVE test samples. The examples cover diverse suspicious regions in complex human-centric scenes. PAVE generally localizes the regions that actually support the authenticity verdict and generates concise explanations describing visible abnormalities inside the selected boxes. In contrast, baseline methods often place boxes on incorrect regions, cover salient objects rather than the supporting evidence, or generate explanations that describe content not visible in the indicated area. These failure modes are consistent with the quantitative gaps in Table~\ref{tab:supp_localization_ours_v_transposed} and Table~\ref{tab:supp_explanation_ours_v_transposed}, and they illustrate why explicit region--explanation alignment is necessary for auditable detection.


\end{document}